\documentclass[preprint,times,English]{elsarticle}

\usepackage{lineno,hyperref,txfonts, geometry,  graphicx, fleqn}
\usepackage{setspace}
\usepackage[T1]{fontenc}
\usepackage{amsmath}
\usepackage{diagbox}
\usepackage{longtable}
\usepackage{multirow}
\usepackage{threeparttable}
\usepackage{graphicx}
\usepackage{subfigure}
\usepackage{overpic}
\usepackage{color}
\usepackage{makecell}
\usepackage{amssymb}
\usepackage{pifont}

\usepackage{amsfonts,amssymb}
\usepackage{mathrsfs}
\usepackage{booktabs}
\graphicspath{{fig/}}
\usepackage{caption}
\usepackage{bm}
\usepackage{upgreek}
\usepackage{float}
\usepackage{enumerate}
\usepackage{pifont}
\usepackage{pdflscape}
\usepackage{array}
\usepackage{appendix}
\usepackage{epstopdf}
\usepackage{overpic}
\usepackage{rotating}
\usepackage{color}
\definecolor{purple}{RGB}{128,0,128}
\usepackage{ulem}
\usepackage{hyperref}  

\usepackage[linesnumbered,ruled,vlined]{algorithm2e}
\usepackage{algpseudocode}
\usepackage{amsmath}

\begin{document}

\begin{frontmatter}

\title{Improving Pseudo Labels With Intra-Class Similarity for Unsupervised Domain Adaptation}

\author{Jie Wang}
\ead{wangjie2017@mail.nwpu.edu.cn}

\author{Xiao-Lei Zhang}
\ead{xiaolei.zhang@nwpu.edu.cn}

\cortext[cor1]{ Corresponding author: Xiao-Lei Zhang}
\address{Center of Intelligent Acoustics and Immersive Communications (CIAIC) and the School of Marine Science and Technology, \\ Northwestern Polytechnical University, Xi'an, Shaanxi 710072, China.}

\begin{abstract}
Unsupervised domain adaptation (UDA) transfers knowledge from a label-rich source domain to a different but related fully-unlabeled target domain. To address the problem
of domain shift, more and more UDA methods adopt pseudo labels of the target samples to improve the generalization ability on the target domain. However, inaccurate pseudo labels of the target samples may yield suboptimal performance with error accumulation during the optimization process. Moreover, once the pseudo labels are generated, how to remedy the generated pseudo labels is far from explored. In this paper, we propose a novel approach to improve the accuracy of the pseudo labels in the target domain. It first generates coarse pseudo labels by a conventional UDA method. Then, it iteratively exploits the intra-class similarity of the target samples for improving the generated coarse pseudo labels, and aligns the source and target domains with the improved pseudo labels. The accuracy improvement of the pseudo labels is made by first deleting dissimilar samples, and then using spanning trees to eliminate the samples with the wrong pseudo labels in the intra-class samples. We have applied the proposed approach to several conventional UDA methods as an additional term.
Experimental results demonstrate that the proposed method can boost the accuracy of the pseudo labels and further lead to more discriminative and domain invariant features than the conventional baselines.
\end{abstract}


\end{frontmatter}

\section{Introduction}

It is known that machine learning benefits from manually-labeled data. However, manual labeling is often time-consuming and laboring intensive. Moreover, it is even unlikely to obtain sufficient manual labels in some scenarios. How to address the insufficient labeling problem is a key task. One of the approaches to address the problem is domain adaptation, which aims to transfer knowledge from a label-rich source domain to a different but related target domain \cite{survey,survey20}. Based on whether the target domain is human labeled, domain adaptation can be divided into two categories \cite{Tclass}, which are semi-supervised domain adaptation \cite{semi1} and unsupervised domain adaptation \cite{Un1}. In this paper, we focus
on unsupervised domain adaptation (UDA) where the target domain does not have manual labels. It is not only challenging but also finds its applications in many real-world scenarios.

In the past decades, UDA has been widely studied \cite{ARC_L,long,LPTJ,CSPP,HDA}. The most common approach is to find a common subspace in which the data distributions of the source domain and target domain are similar. The first issue on learning the subspace is to define a suitable distribution divergence measurement between the two domains. A common measurement is the maximum mean discrepancy (MMD) \cite{MMD1,MMD2,MMDL,MMD33}.
By minimizing the distribution divergence as a regularizer, a common subspace could be found. For example, \cite{TCA} learns a domain-invariant projection and meanwhile minimizes the marginal distribution divergences between the source domain and the target domain.

Recently, a new branch of the UDA research is to use the pseudo labels of the data in the target domain to align the distributions of the source and target domains \cite{DMEA,DICD,SLPP}, where the pseudo labels are usually obtained by a classifier trained on the source domain. For example, following \cite{TCA}, the work \cite{JDA} further reduces the marginal distribution divergence and conditional distribution divergence between the source domain and the target domain iteratively with the pseudo labels of the target domain. \cite{DICD} learns both domain invariant and class discriminative features with the pseudo labels. \cite{LPTJ} preserves the neighborhood relationship of samples and improves robustness against outliers by supervised locality preserving projection \cite{LPP} with the pseudo labels. \cite{reMMD} proposes a discriminative MMD to mitigate the degradation of feature discriminability incurred by MMD.

\begin{figure}[t]
    \centering
    \includegraphics[width=0.75\textwidth]{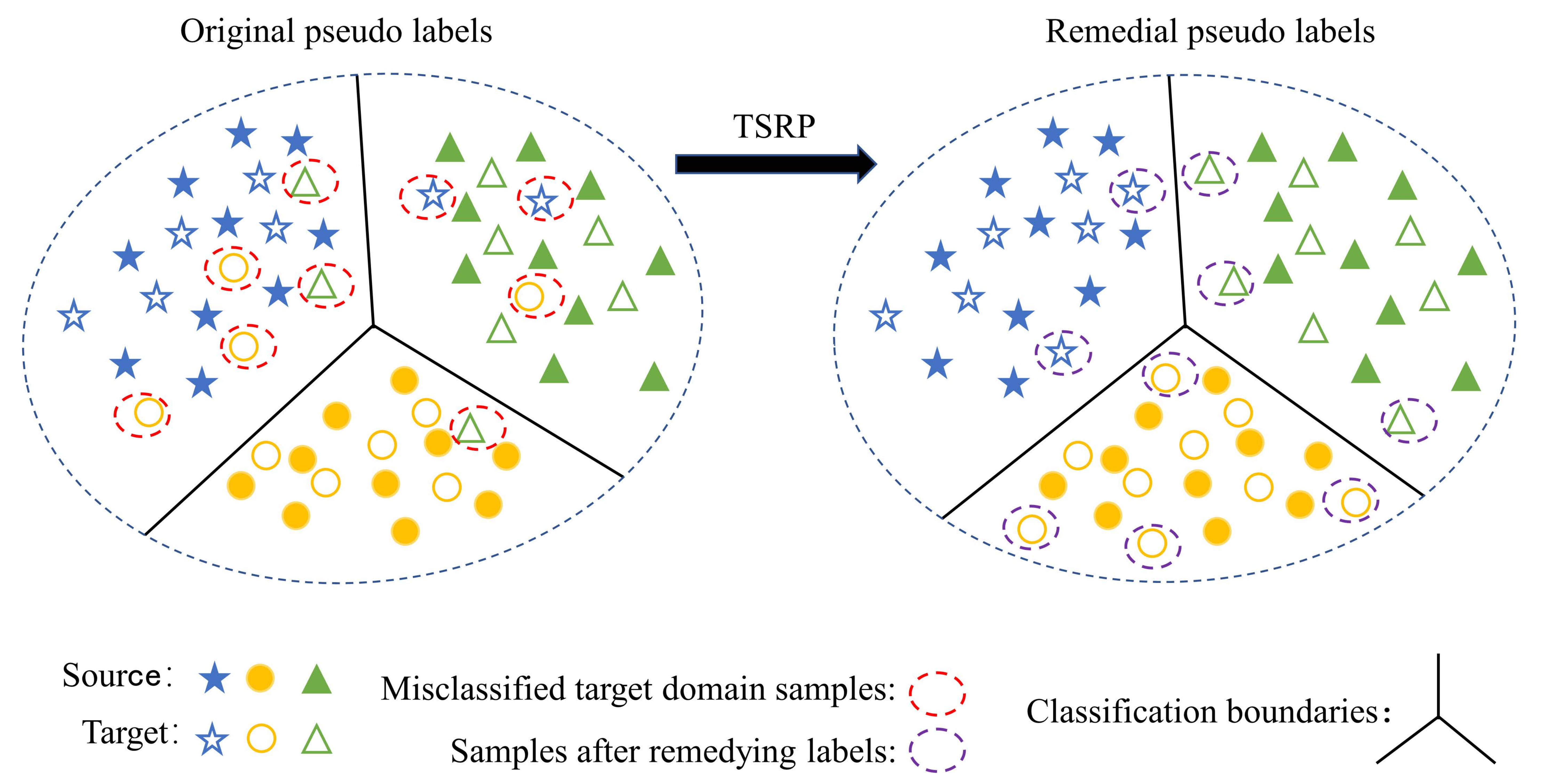}\\
    \caption{Motivation of the proposed TSRP. Most related works focus on learning domain invariance features while ignoring the intra-class similarity between target samples. On the contrary, TSRP aims to explore the intra-class similarity between the samples in the target domain to remedy pseudo labels, which in turn leads to better domain invariance features.}\label{fig.motiation}
\end{figure}

Although the pseudo label generation approaches have made significant contribution to UDA, there are still two issues far from explored.
First, the pseudo labels are mainly obtained by a good alignment between the source domain and the target domain, while the effect of the accuracy of the pseudo labels on performance is not studied deeply. When the pseudo labels are generated by a classifier trained on the source domain, which is the common way, there may be some incorrect pseudo labels in each optimization iteration as shown in Fig. \ref{fig.motiation}. Due to the cumulation of the errors, the incorrect pseudo labels can greatly affect the final performance. Second, most of the methods mainly focus on mining the source domain to improve the accuracy of the pseudo labels in the target domain, however, to our knowledge, the intrinsic relationship between the samples in the target domain seems unexplored yet.


To address the aforementioned two issues, in this paper, we propose to mine the \textit{target domain intra-class similarity to remedy the pseudo labels} (TSRP) in the target domain for improving the accuracy of the pseudo labels. A core idea of TSRP is to \textit{use target similarity to pick pseudo labels with high confidence} (UTSP) by spanning trees \cite{spanning}. Then, the selected highly-confident pseudo-labeled samples as well as the source data are used to train a strong classifier. The strong classifier is used to correct part of the the wrongly-labeled target samples that have low-confident pseudo labels. We call it the \textit{remedial process of the pseudo labels}. Generally, our method can be integrated into any methods that generate the pseudo labels of the target domain by using the classifiers trained on the source domain.

Our contribution is summarized as follows:
\begin{itemize}
\item We propose TSRP to improve the accuracy of the pseudo labels in the target domain. TSRP iteratively exploits the intra-class similarity of the samples in the target domain for improving the generated coarse pseudo labels, and aligns the source and target domains with the improved pseudo labels.

\item We propose UTSP to select highly-confident pseudo-labeled samples. UTSP first deletes dissimilar samples, and then uses spanning trees to eliminate the samples with the wrong pseudo labels in the intra-class samples. 

\item We extended four UDA algorithms \cite{DICD,JDA,NN,BDA} with TSRP, and evaluated the effectiveness of TSRP by comparing the UDA algorithms with their TSRP extensions. Experimental results on several benchmark datasets show that TSRP can be used as a term of the UDA methods for improving their generalization ability. Moreover, we have compared the proposed ``DICD \cite{DICD} +TSRP'' algorithm with a number of representative UDA algorithms \cite{TCA,PCA,geodestic,TJM,TSL,DTSL,CDML,RTML}. Experimental results show that the integrated method behaves better than the comparison methods.
\end{itemize}

The remainder of this paper is organized as follows. In Section \ref{sec:related}, we review some related work. In Section \ref{sec:method}, we propose TSRP to improve the accuracy of pseudo labels. The experimental results are reported in Section \ref{sec:exp}. Finally, we conclude this paper in Section \ref{sec:conclusion}.

\section{Related work}\label{sec:related}   

Early works on UDA aim to align the marginal distributions of the source and target domains\cite{geodestic,backpropagation}. Due to the lack of labeled target samples, even though the marginal distributions are perfectly aligned, there is no guarantee that a good classification result will be produced since that the conditional distribution of the target domain may be misaligned with that of the source domain. To overcome this issue, many UDA methods resort to pseudo labels of the target domain \cite{JDA,CAN,fisher}. If the pseudo labels of the target samples can be properly obtained, then supervised learning can be applied to train a good classifier. There are two strategies to generate the pseudo labels---hard labeling \cite{JDA,GSA,DICD} and soft labeling \cite{muladver}. Because the accuracy of the pseudo labels plays an important role to the quality of the classifier, we summarize some pseudo label generation and selection methods that focus on improving the accuracy of the pseudo labels as follows.

In \cite{triclass}, Saito \textit{et al.} use three asymmetric classifiers to improve the accuracy of the pseudo labels, where two of the classifiers are used to select confident pseudo labels, and the third one aims to learn a discriminative data representation for the target domain. \cite{SLPP} explores the structural information of the target domain by structured prediction, and combines the nearest class prototype and structured prediction to promote the accuracy of pseudo labels. \cite{CMMS} regards the samples of the same cluster in the target domain as a whole rather than individuals. It assigns pseudo labels to the target cluster by class centroid matching. \cite{easyhard} proposed an easy-to-hard strategy which divides target samples into three categories, namely easy samples, hard samples and incorrect-easy samples. It tends to generate pseudo labels for easy samples and tries to avoid hard samples. The easy-to-hard strategy may be biased to easy classes. To address this issue, a confidence-aware pseudo label selection strategy was proposed in \cite{unifying}. It selects samples from each class independently by the probability of pseudo labels. In \cite{easyhard,unifying}, they use the distances from the target samples to the centers of the source samples as the selection criterion to select highly confident pseudo labels. In \cite{SLPP,CMMS}, they iteratively generate-confident pseudo labels. However, they do not consider how to correct the falsely generated pseudo labels.
\begin{figure*}[t]
    \centering
    \includegraphics[width=0.8\textwidth]{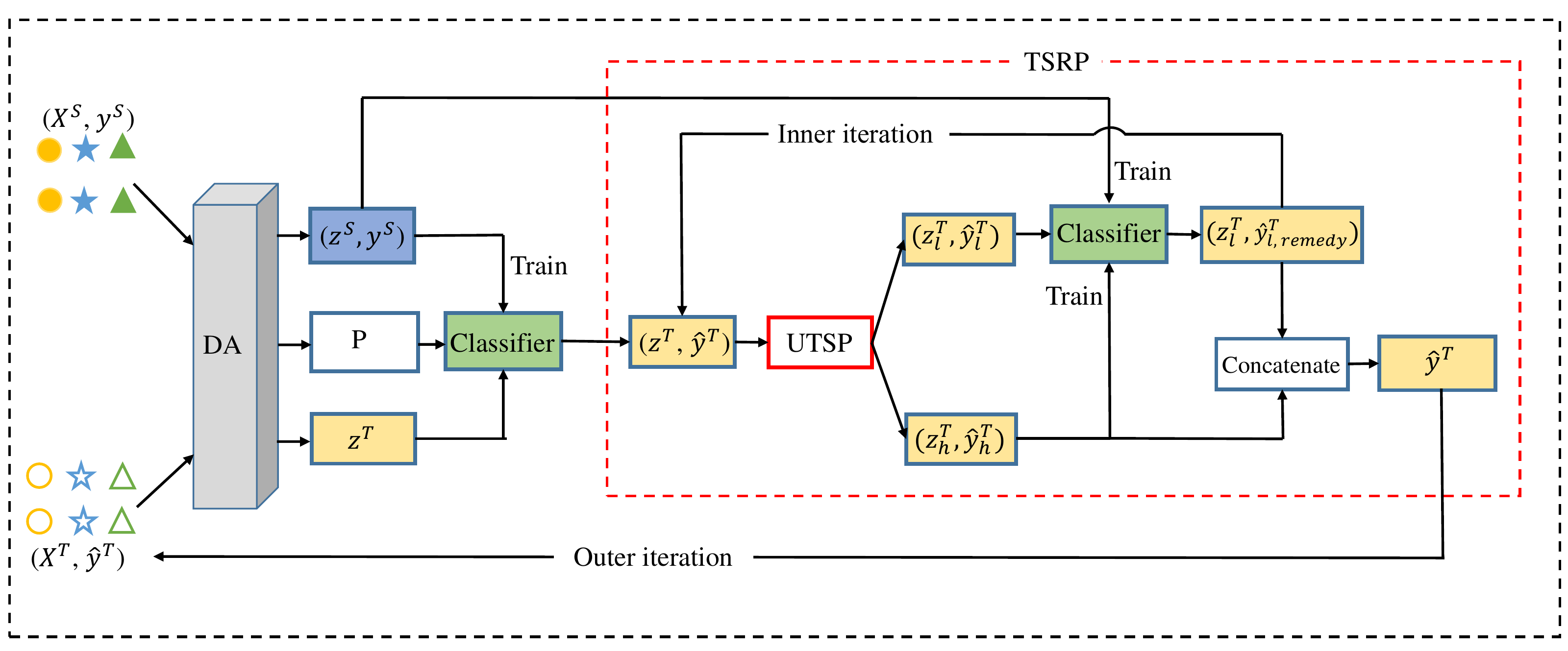}\\
    \caption{Architecture of the proposed framework with TSRP. It consists of the following four successive steps. First, an unsupervised domain adaptation (UDA) method is used to learn a domain-invariant feature. Then, a weak classifier is trained to obtain the pseudo labels of the target samples. Third, UTSP is proposed to pick the target samples with highly-confident pseudo labels. Finally, a strong classifier is trained with the source domain sample and the target samples with the highly-confident pseudo labels, which is used to remedy the pseudo labels.}\label{fig.TSRP}
\end{figure*}
Different from the above methods, in this paper, we propose to correct the falsely generated pseudo labels by exploring the intra-class similarity in the target domain.



\section{Method}\label{sec:method}  

In the following subsections, we give the formulation of the problem and our motivation, and describe the proposed method in detail.

\subsection{Framework}

A UDA problem is formulated as follows. Given a manually labeled source domain $\mathcal{D}^{S}=\left\{\left(\mathbf{x}_{i}^{S}, y_{i}^{{S}}\right)\right\}_{i=1}^{n_{s}}=\left\{\mathbf{X}^{S},\mathbf{y}^{S}\right\}$, and an unlabeled target domain $\mathcal{D}^{T}=\{\mathbf{x}_{j}^{T} \}_{j=1}^{n_{t}}=\left\{\mathbf{X}^{T}\right\}$, where $\mathbf{x}^{S}$ and $\mathbf{x}^{T}$ represent $m$-dimensional feature vectors of the samples in the source domain and target domain respectively, $\mathbf{y}^{{S}}$ is the manual label in the source domain, $n_{s}$ and $n_{t}$ are the number of the source and target samples respectively. The goal of UDA is to predict the labels of $\mathcal{D}^{T}$. In this paper, we focus on the problem that the source and target domains share the same object classes. Suppose there are $C$ classes in both domains.

As summarized in Section \ref{sec:related}, many UDA approaches focus on obtaining a good alignment between the source and target domains to generate pseudo labels, leaving the negative effect of the falsely generated pseudo labels unsolved. Because the inaccurate pseudo labels could result in catastrophic error accumulation during the learning process \cite{SLPP}, intuitively, if we could increase the accuracy of the pseudo labels, then we may get a better alignment between the source domain and the target domain.

In this paper, we propose to steadily improve the accuracy of the pseudo labels in the target domain by iteratively exploiting the intra-class similarity between the target samples. An shown in Fig. \ref{fig.TSRP}. for each iteration, the proposed method runs the following two steps in sequence. First, it employs a traditional UDA method to generate the crude pseudo labels of the target samples for the current iteration given a set of improved pseudo labels from the previous iteration, see Section \ref{subsec:UDA} for the details. Then, it uses TSRP to improve the accuracy of the pseudo labels, see Section \ref{subsec:TSRP} for the details where a key component named UTSP is presented in Section \ref{subsec:UTSP}. 
The overall framework with the TSRP module is summarized in Algorithm \ref{algorithm1}.

\normalem
\begin{algorithm}[t]
\caption{Proposed framework with TSRP module.}\label{algorithm1}
\KwIn{Labeled source samples, $\left\{\mathbf{X}^{S},\mathbf{y}^{S}\right\}=\left\{\left(\mathbf{x}_{i}^{S}, {y}_{i}^{S}\right)\right\}_{i=1}^{n_{s}}$; Target samples, $\left\{\mathbf{X}^{T}\right\}=\{\mathbf{x}_{j}^{T}\}_{j=1}^{n_{t}}$; Trust parameter: $\rho$;}
\KwOut{Remedial pseudo labels of the target domain $\mathbf{\hat{y}}^{T}$;}
\While{not converged}  {
Get the projection matrix $\mathbf{P}$ by a UDA method with $\left\{\mathbf{X}^{S}, \mathbf{y}^{S}\right\}$ and $\left\{\mathbf{X}^{T}, \hat{\mathbf{y}}^{T}\right\}$ \;
$\mathbf{Z}^{S} \leftarrow \mathbf{P}\mathbf{X}^{S}$\;
$\mathbf{Z}^{T} \leftarrow \mathbf{P}\mathbf{X}^{T}$\;
Train the classifier $f(\cdot)$ with $\left\{\mathbf{Z}^{S}, \mathbf{y}^{S}\right\}$\;
$\mathbf{\hat{y}}^{T} \leftarrow f(\mathbf{Z}^{T})$\;
//~\textbf{TSRP} start:
\par
\While{not converged}  {
$(\{\mathbf{Z}^T_h, \hat{\mathbf{y}}^T_h\},\{\mathbf{Z}^T_l, \hat{\mathbf{y}}^T_l\})\leftarrow \mathrm{UTSP}(\{\mathbf{Z}^T, \hat{\mathbf{y}}^T\})$\;
Train a strong classifier $f_{\mathrm{strong}}$ with $(\mathbf{Z}^{S},\mathbf{y}^{S})$ and $(\mathbf{Z}_{h}^{T}, \hat{\mathbf{y}}_{h}^{T})$ \;
Get the remedial pseudo labels $\mathbf{y}_{l, \mathrm{remedy}}^{T}$ by \eqref{eq.strongerclassifier}\;
$\{\mathbf{Z}^T, \hat{\mathbf{y}}^T\}\leftarrow \{\mathbf{Z}^T_l, \mathbf{\hat{y}}^{T}_{l,\mathrm{remedy}}\}$\;
        }
$\mathbf{\hat{y}}^{T}\leftarrow [\mathbf{\hat{y}}^{T}_h,\mathbf{\hat{y}}^{T}_{l,\mathrm{remedy}}]$\;
//~\textbf{TSRP} end\;
 }

\end{algorithm}
\begin{algorithm}[t]
\caption{UTSP.}\label{algorithm2}
\KwIn{Domain-invariant features $\mathbf{z}^{T}$, its corresponding pseudo labels $\mathbf{\hat{y}}^{T}$ which contain $C_y$ pseudo classes;}
\KwOut{Highly-confident samples $\left\{\mathbf{Z}_{h}^{T}, \hat{\mathbf{y}}_{h}^{T}\right\}$, low-confident samples $\left\{\mathbf{Z}_{l}^{T}, \hat{\mathbf{y}}_{l}^{T}\right\}$;}
\For{$k=1,\ldots, C_y$}{
Calculate intra-class similarity matrix $\mathbf{S}^{k}$ by \eqref{eq.similar}\;
Calculate $\delta$ by \eqref{eq.threshold}\;
Calculate adjacency matrix $\mathbf{M}_{k}$ by \eqref{eq.adjacent}\;
Calculate diagonal matrix $\mathbf{D}^{k}$ by \eqref{eq.diagonal}\;
Pick the root node of the spanning tree, i.e. $\mathbf{z}_{m}^{Tk}$, by \eqref{eq.maxdegree}\;
Get highly-confident and low-confident samples by the spanning tree;
}
\end{algorithm}
\subsection{UDA: Generating crude pseudo labels}\label{subsec:UDA}

 An existing UDA algorithm is employed to learn a projection matrix $\mathbf{P}$ that maps the samples from both domains into a shared latent subspace. A requirement is that $\mathbf{P}$ should be learned in a supervised manner where the remedial pseudo labels of the target samples obtained from the previous iteration of the framework. $\mathbf{\hat{y}}^{T} = [\hat{y}^T_{1},\ldots,\hat{y}^T_{n_t}]$ are treated as the labels. Various advanced UDA algorithms meet this requirement, such as those \cite{DICD,JDA,NN,BDA} employed in the experiments. Then, domain invariant features of the source and target domains are obtained by $\mathbf{z}^{S}_{i} = \mathbf{P}\mathbf{x}^{S}_i$ and
$\mathbf{z}^{T}_j = \mathbf{P}\mathbf{x}^{T}_j$. Finally, a classifier $f(\cdot)$ is trained with $\left\{\mathbf{Z}^{S}, \mathbf{y}^{S}\right\}$ where $\mathbf{Z}^{S} = [\mathbf{z}^S_1,\ldots,\mathbf{z}^S_{n_s}]$. It is then used to classify $\mathbf{Z}^{T} = [\mathbf{z}^{T}_1,\ldots,\mathbf{z}^T_{n_t}]$ into $C_{y}$ classes ($C_{y}\leq C$). The predicted labels of the target samples are denoted as the \textit{crude pseudo labels} $\mathbf{\hat{y}}^{T}=[\hat{y}^{T}_1,\ldots,\hat{y}^{T}_{n_t}]$.

\subsection{TSRP: Using target domain intra-class similarity to remedy pseudo labels}\label{subsec:TSRP}
There may always be some incorrect pseudo labels in $\mathbf{\hat{y}}^{T}$ in each iteration, especially, in the early training stage, therefore, when learning $\mathbf{P}$ with the incorrect pseudo labels, the final performance may not be good due to the cumulative errors in the iterative optimization process. If we could correct part of the incorrect pseudo labels in each iteration, then the performance might be improved steadily. To address this issue, a possible way is to pick the pseudo labels with low confidence, and conduct correction to them with a stronger classifier than the original classifier $f(\cdot)$.

Motivated by the above analysis, TSRP aims to improve the accuracy of the crude pseudo labels by remedying the pseudo labels with low confidence.
Specifically, TSRP first partitions the target samples into two sets, one with highly-confident pseudo labels, denoted as $\{\mathbf{Z}^T_h, \hat{\mathbf{y}}^T_h\}$ and the other one with low confident pseudo labels, denoted as $\{\mathbf{Z}^T_l, \hat{\mathbf{y}}^T_l\}$, by exploring the intra-class similarity in the target domain (see Section \ref{subsec:UTSP}). Then, it trains a strong classifier $f_{\mathrm{strong}}(\cdot)$ with $\{\mathbf{Z}^T_h, \hat{\mathbf{y}}^T_h\}$ and $\{\mathbf{Z}^S, \mathbf{y}^S\}$, and uses the classifier to predict the labels of $\mathbf{Z}^T_l$:
\begin{equation}
\label{eq.strongerclassifier}
\begin{array}{cl}
\hat{\mathbf{y}}^T_{l, \mathrm{remedy}}=f_{\mathrm{strong}}\left(\mathbf{Z}_{l}^{T}\right)
\end{array}
\end{equation}
where $\hat{\mathbf{y}}^T_{l, \mathrm{remedy}}$ is the remedial pseudo labels of the target samples with low confidence. Finally, we take the remedial pseudo labels and the pseudo labels with high confidence together as the target domain pseudo labels to train $\mathbf{P}$ in the next iteration.

\subsection{UTSP: Using target intra-class similarity to pick pseudo labels with high confidence}\label{subsec:UTSP}
\begin{figure}[t]
    \centering
    \includegraphics[width=0.65\textwidth]{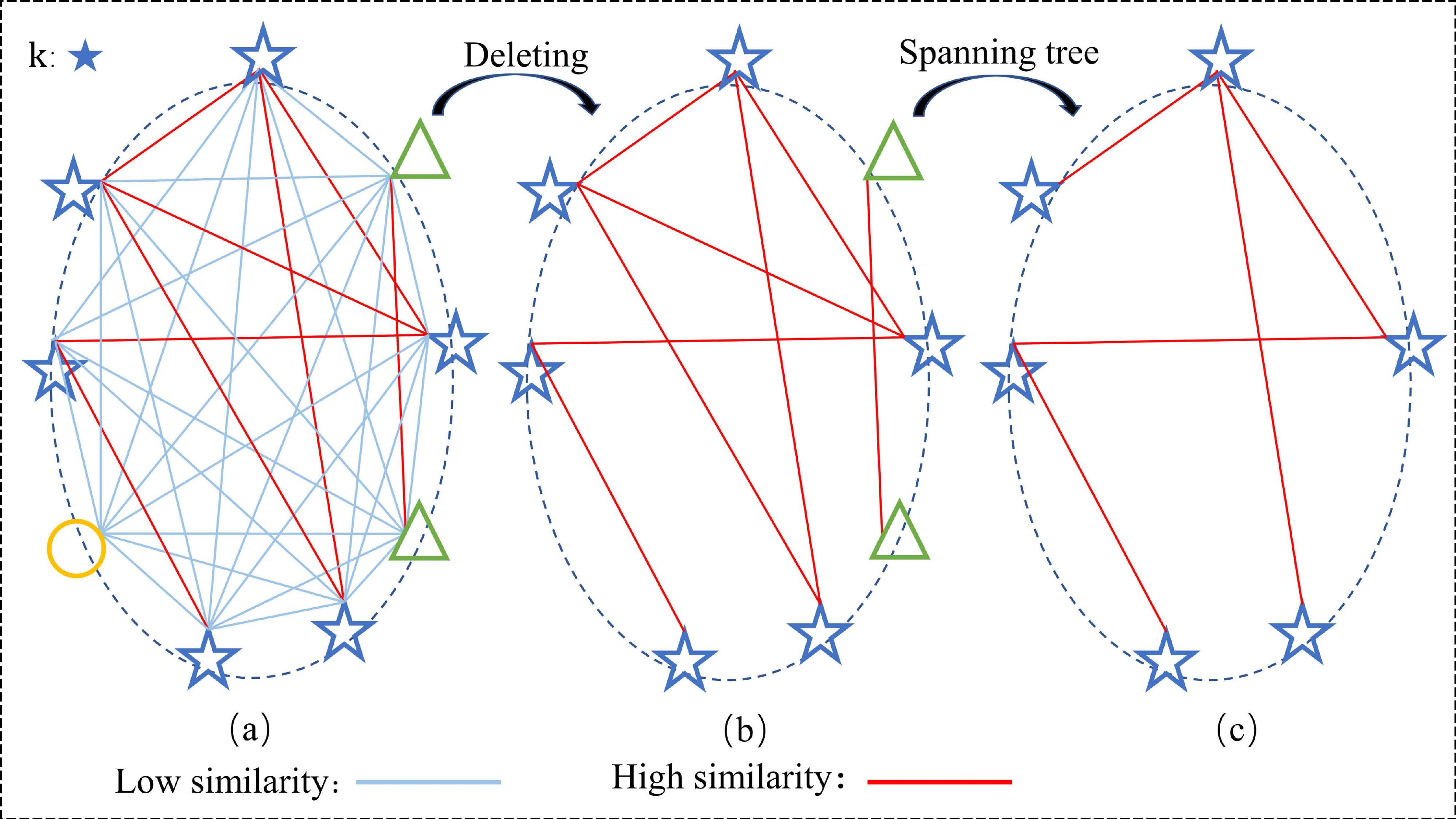}\\
    \caption{ Principle of UTSP. UTSP consists of two steps: (i) \textit{Deleting}, which deletes the samples with low pairwise similarity scores as shown from Fig. (a) to Fig. (b), and (ii) \textit{spanning tree}, which selects samples with highly-confident pseudo labels by spanning trees as shown from Fig. (b) to Fig. (c).}\label{fig.Pick}
\end{figure}

UTSP aims to select the target samples whose pseudo labels are highly-confident. It consists of two steps---\textit{deleting} and \textit{spanning tree}. Its principle is illustrated in Fig. \ref{fig.Pick}. We will present the details of the two steps in the following subsections with a summary in Algorithm \ref{algorithm2}.

\subsubsection{Deleting}
The \textit{deleting} step aims to delete the samples with small similarity for each pseudo class. In the following, we present the \textit{deleting} step for the $k$-th pseudo class, $\forall k=1,2,\ldots,C_{y}$.
It first calculates a pairwise intra-class similarity matrix $\mathbf{S}^{k}$ of the target samples by e.g. cosine similarity:
\begin{equation}
\label{eq.similar}
S_{i,j}^{k}= \left\{
\begin{array}{ll}
0, & i=j \\
\frac{\langle\mathbf{z}_{i}^{Tk}, \mathbf{z}_{j}^{Tk}\rangle}{\|\mathbf{z}_{i}^{Tk}\|\|\mathbf{z}_{j}^{Tk}\|}, & \mbox{otherwise}\\
\end{array}\right.,
\forall k=1,2,\ldots,C_{y}
\end{equation}
where $S_{i,j}^{k}$ denotes the cosine similarity between the $i$-th sample and $j$-th sample of the $k$-th pseudo class, $i,j\in\left\{1,2,\ldots,n_{t}^{k}\right\}$ with $n_t^k$ denoted as the number of samples in the $k$-th pseudo class, $n_t =\sum_{k=1}^{C_{y}}n_t^k$, and $\mathbf{z}^{Tk}$ denotes a target sample belonging to the $k$-th pseudo class. Because $\mathbf{S}^{k}$ is a symmetric matrix, we only keep the upper triangular matrix of $\mathbf{S}^{k}$, denoted as $\mathbf{S}_{\mathrm{upper}}^{k}$. We sort the non-zero elements of $\mathbf{S}_{\mathrm{upper}}^{k}$ in the ascending order:
\begin{equation}
S_{\mathrm{rank}}^{k}=\left\{{S_{\mathrm{rank}}^{k}}_{(1)}, {S_{\mathrm{rank}}^{k}}_{(2)}, \ldots, {S_{\mathrm{rank}}^{k}}_{(n_p)}\right\}
\end{equation}
where $n_p$ is the number of non-zero elements in $\mathbf{S}_{\mathrm{upper}}^{k}$.

In order to select highly-confident pseudo labels in each category, the \textit{deleting} step sets a similarity threshold $\delta$:
\begin{equation}
\label{eq.threshold}
\delta = {S_{\mathrm{rank}}^{k}}_{(\lfloor\rho  n_p\rfloor)}
\end{equation}
where $\rho \in (0,1)$ is a trust parameter. Then, we can obtain a mask matrix $\mathbf{M}^{k}$ as follows:
\begin{equation}
\label{eq.adjacent}
M_{i j}^{k}= \begin{cases}1, & S_{i,j}^{k}\geq \delta \\ 0, & \mathrm { otherwise }\end{cases}
\end{equation}

 If we regard each sample $\mathbf{z}^{Tk}$ as a node, and take $\mathbf{M}^{k}$ as the adjacency matrix of the undirected graph that are composed of the $n_{t}^{k}$ nodes, then we could observe commonly that the samples belonging to the same ground-truth category have high pairwise similarity scores in the pseudo class, on the contrary, the samples belonging to different ground-truth categories have low pairwise similarity scores. Here we regard the nodes with no neighbors as the samples with low-confident pseudo labels. We delete these nodes from the undirected graph. The process is illustrated in Fig. \ref{fig.Pick}a to Fig. \ref{fig.Pick}b.

\subsubsection{Spanning tree}
However, the samples of the $k$-th pseudo class may be mixed with samples from multiple ground-truth categories that are geometrically very similar to each other. If we only select highly-confident pseudo labels by the threshold $\delta$, some misclassified samples may be selected as highly-confident samples after the \textit{deleting} step.

To further refine the samples selected by the \textit{deleting} step, we explore an idea from \textit{spanning forests} \cite{spanning}. Specifically, we first get a diagonal matrix $\mathbf{D}^{k}$ by:
\begin{equation}
\label{eq.diagonal}
D^{k}_{i i}=\sum_{j} M^{k}_{ij}
\end{equation}
The identity of the largest element of $\mathbf{D}^{k}$ can be computed as:
\begin{equation}
\label{eq.maxdegree}
m=\arg\max_i D^{k}_{i i}
\end{equation}
The node with the maximum degree, i.e. $\mathbf{z}_{m}^{Tk}$, represents the most confident sample in the $k$ pseudo class, since that it has the maximum number of neighbors.

Then, we set $\mathbf{z}_{m}^{Tk}$ to the root node of a spanning tree and find its leaf nodes. We regard the samples in the same spanning tree as highly-confident samples of the $k$-th pseudo class, and the samples that are not in the spanning tree as misclassified samples from other ground-truth categories. This process is illustrated in Fig. \ref{fig.Pick}b to Fig. \ref{fig.Pick}c.

By pooling the highly-confident samples throughout all pseudo classes, we finally get the highly-confident set $\{\mathbf{Z}^T_h, \hat{\mathbf{y}}^T_h\}$ and low-confident set $\{\mathbf{Z}^T_l, \hat{\mathbf{y}}^T_l\}$. Note that, in order to avoid a bad solution to class-imbalanced problems where a class with a small number of samples disappears after UTSP, we regard all samples of the pseudo class satisfying $n_{t}^{k} \leq 3$ as highly-confident samples.

\section{Experiments}\label{sec:exp}
In this section, we evaluate the performance of the proposed methods. The source code of TSRP is available at https://github.com/02Bigboy/TSRP.

\subsection{Datasets and cross-domain tasks}
We used common cross-domain datasets \cite{DICD}, including CMU-PIE \cite{PIE}, MNIST \cite{handwritten2}, and USPS \cite{handwritten}, Office \cite{geodestic,JDA}, Caltech256 \cite{Caltech}, and COIL20 \cite{COIL}. The datasets are described in Table \ref{tab:data}. The domain adaptation tasks on the datasets are described as follows.

CMU-PIE is a large face dataset. It consists of more than 40,000 face images from 68 individuals. The face images vary widely due to the variations of the illumination condition, poses, and expressions. In terms of different pose factors, we chose five subsets as in \cite{DICD}: C05 (left pose), C07 (upward pose), C09 (downward pose), C27 (front pose) and C29 (right pose) to construct the cross-domain classification tasks. Following the experimental setting in \cite{DICD}, we randomly selected two subsets as the source domain and target domain respectively for each cross-domain task, which results in 20 cross-domain tasks in total, e.g. ``$\mathrm{C05} \rightarrow \mathrm{C07}$'', ``$\mathrm{C05}\rightarrow\mathrm{C09}$'', ... , ``$\mathrm{C29}\rightarrow\mathrm{C27}$'' in the form of ``source$\rightarrow$target''.

\begin{table}[t]
\centering
\caption{Description of the visual cross-domain datasets in the experiments.}\label{tab:data}
\begin{tabular}{|l||l|c|c|c|}
\hline Dataset & Type & Samples & Classes & Features \\
\hline CMU-PIE & Face & 11554 & 68 & 1024 \\
\hline MNIST & Digit & 2000 & 10 & 256 \\
\hline USPS & Digit & 1800 & 10 & 256 \\
\hline AMAZON (A) & Object & 958 & 10 & $800 / 4096$ \\
\hline CALTECH (C) & Object & 1123 & 10 & $800 / 4096$ \\
\hline DSLR (D) & Object & 157 & 10 & $800 / 4096$ \\
\hline WEBCAM (W) & Object & 295 & 10 & $800 / 4096$ \\
\hline COIL20 & Object & 1440 & 20 & 1024 \\
\hline
\end{tabular}
\end{table}

MNIST-USPS consists of two classical hand written digit image datasets---USPS \cite{handwritten} and MNIST \cite{handwritten2}. To speed up the experimental comparisons as that in \cite{JDA}, we randomly chose 1,800 images from USPS and 2,000 images from MNIST, and rescaled the images to a size of $16 \times 16$, which forms two cross-domain tasks, i.e. ``$\mathrm{MNIST}\rightarrow\mathrm{USPS}$'' and ``$\mathrm{USPS}\rightarrow\mathrm{MNIST}$''.

Office+Caltech dataset is one of the most commonly used datasets for unsupervised domain adaptation. It consists of four domains: Amazon (images downloaded from online merchants), Webcam (low-resolution images by a web camera), DSLR (high-resolution images by a digital SLR camera) and Caltech-256. Here, ten common classes from all four domains were used, which are backpack, bike, calculator, headphone, computer-keyboard, laptop, computer-monitor, computer-mouse, coffee-mug, and video-projector respectively. There are 2,533 images in total with 8 to 151 images per category per domain.
We extracted two kinds of features, which are the 800-dim SURF \cite{geodestic} and 4096-dim DeCAF6 \cite{Decaf} respectively. Similar to \cite{DICD}, we obtained 12 cross-domain tasks for two kinds of features, e.g. ``$\mathrm{A}\rightarrow\mathrm{C}$'', ``$\mathrm{A}\rightarrow\mathrm{D}$'' , ... , ``$\mathrm{W}\rightarrow\mathrm{C}$'' and ``$\mathrm{W}\rightarrow\mathrm{D}$'', by randomly choosing two domains from the data as the source and target respectively.

COIL20 dataset consists of 1,440 grayscale images with 20 objects. Each object has 72 images of size $32 \times 32$ taken at pose intervals of 5 degrees rotating through 360 degrees. Like \cite{DICD}, we split the dataset into 2 subsets: COIL1 and COIL2. COIL1 includes all images at the directions of $0^\circ$ to $85^\circ$ and $180^\circ$ to $265^\circ$. COIL2 contains the images of $90^\circ$ to $175^\circ$ and $270^\circ$ to $355^\circ$. They follow different but related distributions since that COIL1 and COIL2 consist of the same objects with diverse shooting degrees. We randomly chose one as the source domain and the other as the target domain for two cross-domain tasks, e.g. ``$\mathrm{COIL1} \rightarrow \mathrm{COIL2}$'' and ``$\mathrm{COIL2}\rightarrow\mathrm{COIL1}$''.

\subsection{Experimental settings}

 TSRP can be used as a term of many UDA algorithms. In order to verify the effectiveness of TSRP, we integrated it into 4 algorithms, namely 1-nearest neighbor (NN) \cite{NN},  joint distribution adaptation (JDA) \cite{JDA}, balanced distribution
adaptation (BDA) \cite{BDA}, and domain invariant and class discriminative feature learning (DICD) \cite{DICD}. NN is a standard machine learning methods. JDA aligns both marginal distribution and conditional distribution of of the source and target domains. BDA aligns the marginal distribution and conditional distribution with different weights according to different tasks. DICD considers the class discrimination to learn both domain invariant and class discriminative features. We denote the extended methods of the above four UDA algorithms as NN+TSRP, JDA+TSRP, BDA+TSRP, and DICD+TSRP respectively.

Our TSRP approach consists of two hyper-parameters: The trust parameter $\rho$, and the number of inner iterations for TSRP $IT$. We set $IT = 3$ as a fixed parameter for all experiments. We set $\rho$  = 0.9 for the datasets Office+Caltech-256 (SURF) and CMU-PIE, and $\rho$ = 0.85 for the other datasets. The iteration number for JDA, BDA, DICD was set to $T = 10$. For a fair comparison, the parameters of the extended methods and the original methods were set the same. The classification accuracy on the target domain was used as the evaluation metric.

\subsection{Experimental results}
In this section, we compared the four UDA algorithms, i.e. NN, JDA, BDA, and DICD,  with their TSRP extensions on the visual cross-domain tasks.

Table \ref{tab:PIE} shows the classification performance on the CMU-PIE dataset. From the table, we can see that, after incorporating TSRP, the performance of JDA, BDA, and DICD has been significantly improved. To be specific, DICD+TSRP, BDA+TSRP, and JDA+TSRP achieve $4.75\%$, $2.52\%$, $2.05\%$ absolute improvement respectively over DICD, BDA, and JDA in terms of average accuracy. It is worthy noting that, compared to DICD, DICD+TSRP obtains the best results on all cross-domain tasks of PIE. BDA+TSRP achieves the best results in 17 tasks out of all 20 tasks, compared to BDA. JDA+TSRP achieves the best results in 15 tasks compared to JDA. In addition, we have also observed that the effect of NN+TSRP is worse than that of NN. It may be caused by that the original data of PIE in the source domain and target domain are quite different. Therefore, when NN is applied to the original data directly, the incorrectly generated pseudo labels may be the majority. Eventually, the majority is further enhanced after TSRP is applied.

Table \ref{tab:surf} lists the classification accuracy of the comparison methods on the Office+Caltech-256 (SURF features) dataset. From the table, we see that, after combining with TSRP, all of the four UDA algorithms have been improved. Specifically, among the 12 tasks, DICD+TSRP, BDA+TSRP, JDA+TSRP, and NN+TSRP outperforms their original counterparts in 9, 8, 7, and 6 tasks respectively.

\renewcommand\arraystretch{1}
\begin{table*}[t]
\centering
\caption{Average classification accuracy (\%) of the comparison methods on the target domains of the CMU-PIE tasks.}
\label{tab:PIE}
\scalebox{0.8}{
\begin{tabular}{|c||c|c|c|c|c|c|c|c|}
\hline
\diagbox{Tasks}{Methods} & NN     & NN+TSRP & JDA   & JDA+TSRP & BDA   & BDA+TSRP & DICD  & DICD+TSRP \\ \hline
C05 $\rightarrow$ C07       & 26.09  & 23.51   & 58.81 & $\mathbf{63.41}$    & 58.81 & $\mathbf{64.52}$    & 72.99 & $\mathbf{74.52}$     \\ \hline
C05 $\rightarrow$ C09       & 26.59  & 22.92   & 54.23 & $\mathbf{56.92}$    & 57.11 & $\mathbf{58.70}$    & 72.00 & $\mathbf{76.16}$     \\ \hline
C05 $\rightarrow$ C27       & 30.67  & 27.31   & 84.50 & 81.98               & 84.50 & 82.31               & 92.22 & $\mathbf{96.52}$     \\ \hline
C05 $\rightarrow$ C29       & 16.67  & 14.64   & 49.75 & $\mathbf{54.47}$    & 49.94 & $\mathbf{57.41}$    & 66.85 & $\mathbf{71.32}$     \\ \hline
C07 $\rightarrow$ C05       & 24.49  & 24.10   & 57.62 & $\mathbf{63.00}$    & 57.77 & $\mathbf{64.02}$    & 69.93 & $\mathbf{73.20}$     \\ \hline
C07 $\rightarrow$ C09       & 46.63  & 41.30   & 62.93 & $\mathbf{62.93}$    & 62.93 & $\mathbf{63.60}$    & 65.87 & $\mathbf{67.83}$     \\ \hline
C07 $\rightarrow$ C27       & 54.07  & 52.42   & 75.82 & 75.52               & 76.06 & $\mathbf{77.20}$    & 85.25 & $\mathbf{86.42}$     \\ \hline
C07 $\rightarrow$ C29       & 26.53  & 23.28   & 39.89 & $\mathbf{46.69}$    & 42.03 & $\mathbf{46.69}$    & 48.71 & $\mathbf{56.43}$     \\ \hline
C09 $\rightarrow$ C05       & 21. 37 & \textbf{21.76}   & 50.96 & 48.08               & 52.76 & 51.38               & 69.36 & $\mathbf{70.14}$     \\ \hline
C09 $\rightarrow$ C07       & 41.01  & 34.56   & 57.95 & $\mathbf{58.13}$    & 57.95 & $\mathbf{58.99}$    & 65.44 & $\mathbf{73.05}$     \\ \hline
C09 $\rightarrow$ C27       & 46.53  & 46.14   & 68.45 & $\mathbf{69.45}$    & 68.88 & $\mathbf{70.68}$    & 83.39 & $\mathbf{94.26}$     \\ \hline
C09 $\rightarrow$ C29       & 26.23  & 23.65   & 39.95 & $\mathbf{46.14}$    & 42.65 & $\mathbf{46.81}$    & 61.40 & $\mathbf{66.48}$     \\ \hline
C27 $\rightarrow$ C05       & 32.95  & 30.58   & 80.58 & $\mathbf{81.54}$    & 80.70 & $\mathbf{81.99}$    & 93.13 & $\mathbf{94.72}$     \\ \hline
C27 $\rightarrow$ C07       & 62.68  & 59.30   & 82.63 & 82.01               & 83.18 & $\mathbf{83.92}$    & 90.12 & $\mathbf{92.88}$     \\ \hline
C27 $\rightarrow$ C09       & 73.22  & 72.30   & 87.25 & 86.52               & 87.32 & 87.13               & 88.97 & $\mathbf{90.26}$     \\ \hline
C27 $\rightarrow$ C29       & 37.19  & 33.58   & 54.66 & $\mathbf{57.54}$    & 55.64 & $\mathbf{61.15}$    & 75.61 & $\mathbf{79.11}$     \\ \hline
C29 $\rightarrow$ C05       & 18.49  & 18.40   & 46.46 & $\mathbf{56.66}$    & 50.99 & $\mathbf{57.47}$    & 62.88 & $\mathbf{73.68}$     \\ \hline
C29 $\rightarrow$ C07       & 24.19  & 21.42   & 42.05 & $\mathbf{44.38}$    & 45.92 & $\mathbf{46.59}$    & 57.03 & $\mathbf{65.81}$     \\ \hline
C29 $\rightarrow$ C09       & 28.31  & 27.14   & 53.31 & 50.86               & 53.25 & $\mathbf{55.76}$    & 65.87 & $\mathbf{70.10}$     \\ \hline
C29 $\rightarrow$ C27       & 31.24  & \textbf{31.30}   & 57.01 & $\mathbf{59.63}$    & 57.28 & $\mathbf{59.63}$    & 74.77 & $\mathbf{83.81}$     \\ \hline
Average accuracy                     & 35.46  & 32.48   & 60.24 & $\mathbf{62.29}$    & 61.28 & $\mathbf{63.80}$    & 73.09 & $\mathbf{77.84}$     \\ \hline
Average improvement               &        & -2.98   &       & $\mathbf{2.05}$     &       & $\mathbf{2.52} $    &       & $\mathbf{4.75} $      \\ \hline
Relative improvement              &        & -3.49   &       & $\mathbf{5.16}$     &       & $\mathbf{6.5}  $    &       & $\mathbf{17.65}$      \\ \hline
\end{tabular}
}
\end{table*}

\renewcommand\arraystretch{1}
\begin{table*}[t]
\centering
\caption{Classification accuracy (\%) on the Office+Caltech-256 (surf features) tasks, where A = AMAZON, C = CALTECH, D = DSLR and W = WEBCAM}
\label{tab:surf}
\scalebox{0.8}{
\begin{tabular}{|c||c|c|c|c|c|c|c|c|}
\hline
\diagbox{Tasks}{Methods} & NN    & NN+TSRP        & JDA   & JDA+TSRP       & BDA   & BDA+TSRP       & DICD  & DICD+TSRP      \\ \hline
C $\rightarrow$ A (SURF)      & 23.70 & 23.49          & 44.78 & \textbf{46.45} & 46.14 & \textbf{48.33} & 47.29 & \textbf{47.81} \\ \hline
C $\rightarrow$ W (SURF)      & 25.76 & 24.75          & 41.69 & \textbf{46.10} & 41.69 & \textbf{47.46} & 46.44 & \textbf{50.85} \\ \hline
C $\rightarrow$ D (SURF)      & 25.48 & 24.84          & 45.22 & \textbf{49.04} & 47.13 & \textbf{49.04} & 49.68 & \textbf{50.96} \\ \hline
A $\rightarrow$ C (SURF)      & 26.00 & \textbf{26.63} & 39.36 & \textbf{39.63} & 40.61 & 39.72          & 42.39 & 41.76          \\ \hline
A $\rightarrow$ W (SURF)      & 29.83 & \textbf{30.17} & 37.97 & \textbf{43.39} & 40.00 & 39.72          & 45.08 & \textbf{49.15} \\ \hline
A $\rightarrow$ D (SURF)      & 25.48 & \textbf{26.75} & 39.49 & 31.85          & 40.13 & 38.85          & 38.85 & \textbf{42.04} \\ \hline
w $\rightarrow$ C (SURF)      & 19.86 & 18.25          & 31.17 & \textbf{31.52} & 32.06 & \textbf{33.04} & 33.57 & 32.95          \\ \hline
w $\rightarrow$ A (SURF)      & 22.96 & 21.92          & 32.78 & 30.48          & 32.99 & 32.15          & 34.13 & 31.94          \\ \hline
w $\rightarrow$ D (SURF)      & 59.24 & \textbf{59.87} & 89.17 & \textbf{89.81} & 89.17 & \textbf{90.45} & 89.81 & \textbf{89.81} \\ \hline
D $\rightarrow$ C (SURF)      & 26.27 & 26.09          & 31.52 & 31.43          & 33.39 & \textbf{33.57} & 34.64 & \textbf{37.04} \\ \hline
D $\rightarrow$ A (SURF)      & 28.50 & \textbf{29.33} & 33.09 & 32.78          & 33.72 & \textbf{34.03} & 34.45 & \textbf{35.28} \\ \hline
D $\rightarrow$ W (SURF)      & 63.39 & \textbf{65.08} & 89.49 & 88.47          & 89.49 & \textbf{90.51} & 91.19 & \textbf{91.19} \\ \hline
Average accuracy                      & 31.37 & \textbf{31.43} & 46.31 & \textbf{46.75} & 47.21 & \textbf{48.07} & 48.96 & \textbf{50.06} \\ \hline
Average improvement                 &       & \textbf{0.06}  &       & \textbf{0.44}  &       & \textbf{0.86}  &       & \textbf{1.10}  \\ \hline
Relative improvement                &       & \textbf{0.09}  &       & \textbf{0.82}  &       & \textbf{1.63}  &       & \textbf{2.16}  \\ \hline
\end{tabular}
}
\end{table*}

Tables \ref{tab:CandM} and \ref{tab:C} list the classification accuracy on MNIST+USPS and COIL20 datasets. From the tables, we observe that JDA+TSRP, BDA+TSRP and DICD+TSRP outperform their original counterparts on all tasks. Particularly, DICD+TSRP achieves a relative improvement of $47.67\%$ over DICD on COIL20.

\renewcommand\arraystretch{1}
\begin{table*}[t]
\centering
\caption{Classification accuracy (\%) on the MNIST+USPS tasks.}
\label{tab:CandM}
\scalebox{0.8}{
\begin{tabular}{|c||c|c|c|c|c|c|c|c|}
\hline
\diagbox{Tasks}{Methods}  & NN    & NN+TSRP & JDA    & JDA+TSRP & BDA   & BDA+TSRP & DICD     & DICD+TSRP \\\hline
USPS $\rightarrow$ MNIST     & 35.85 & 35.30            & 59.65 & \textbf{62.40}    & 60.05 & \textbf{62.40}    & 61.50    & \textbf{67.55}     \\\hline
MNIST $\rightarrow$ USPS     & 64.44 & \textbf{70.72}   & 67.28 & \textbf{72.44}    & 69.89 & \textbf{74.06}    & 73.28  & \textbf{73.39}     \\\hline
Average accuracy                     & 50.15 & \textbf{53.01}   & 63.46 & \textbf{67.42}    & 64.97 & \textbf{68.23}    & 67.39    & \textbf{70.47}     \\\hline
Average   improvement              &       & \textbf{2.86}    &       & \textbf{3.96}     &       & \textbf{3.26}     &          & \textbf{3.08}      \\\hline
Relative improvement               &       & \textbf{5.74}    &       & \textbf{10.84}    &       & \textbf{9.30}     &          & \textbf{9.45}  \\\hline
\end{tabular}
}
\end{table*}

\renewcommand\arraystretch{1}
\begin{table*}[t]
\centering
\caption{Classification accuracy (\%) on the COIL20 tasks.}
\label{tab:C}
\scalebox{0.8}{
\begin{tabular}{|c||c|c|c|c|c|c|c|c|}
\hline
\diagbox{Tasks}{Methods}  & NN    & NN+TSRP & JDA    & JDA+TSRP & BDA   & BDA+TSRP & DICD     & DICD+TSRP \\\hline
COIL1 $\rightarrow$ COIL2    & 84.72 & \textbf{88.75}   & 93.75 & \textbf{95.14}    & 93.89 & \textbf{96.53}    & 94.58    & \textbf{95.69}     \\\hline
COIL2 $\rightarrow$ COIL1    & 83.33 & 83.19            & 92.64 & \textbf{94.86}    & 93.33 & \textbf{95.56}    & 93.47    & \textbf{98.06}     \\\hline
Average accuracy                    & 84.03 & \textbf{85.97}   & 93.19 & \textbf{95.00}    & 93.61 & \textbf{96.04}    & 94.03    & \textbf{96.88}     \\\hline
Average  improvement               &       & \textbf{1.94}    &       & \textbf{1.81}     &       & \textbf{2.43}     &          & \textbf{2.85}      \\\hline
Relative improvement               &       & \textbf{12.17}   &       & \textbf{26.53}    &       & \textbf{38.04}    &          & \textbf{47.67}     \\\hline
\end{tabular}
}
\end{table*}

\begin{figure}[t]
    \centering
    \includegraphics[width=0.75\textwidth]{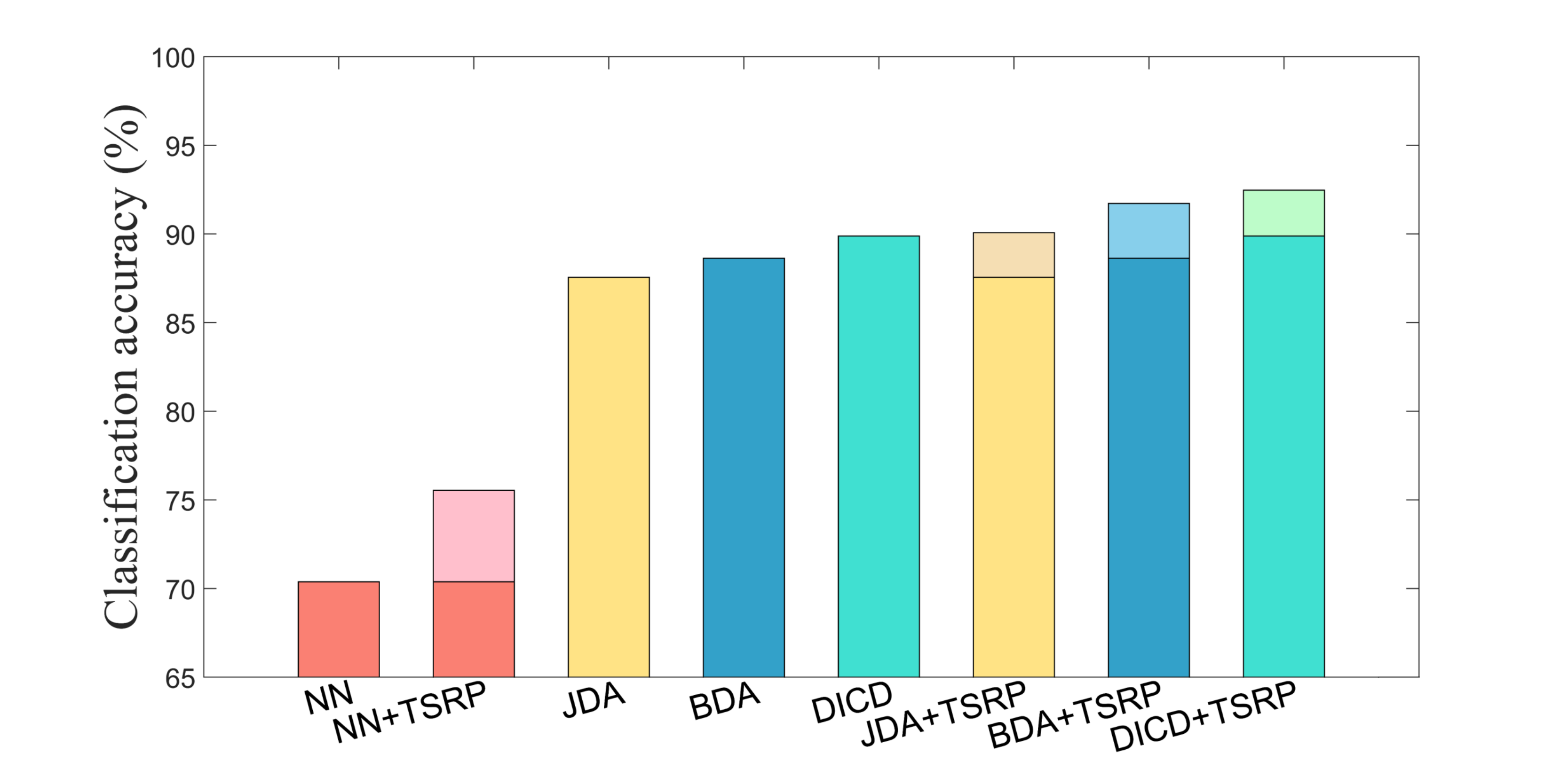}\\
    \caption{Performance summary of the domain adaptation baselines and their TSRP extensions on Office+Caltech-256 in terms of average classification accuracy.}\label{fig.MeTSRP}
\end{figure}

Table \ref{tab:decaf} shows the results on the Office+Caltech-256 (DECAF6 features) dataset. We see from the table that the average accuracy of NN+TSRP, JDA+TSRP, BDA+TSRP and DICD+TSRP is $4.19\%$, $2.52\%$, $ 2.58\%$, $2.59\%$ higher than that of NN, JDA, BDA and DICD respectively. Particularly, BDA+TSRP and DICD+TSRP outperform their original counterparts on all tasks. Both JDA+TSRP and NN+TSRP outperform their counterparts in 11 out of 12 tasks. In order to better show the advantage of TSRP, we further draw the average accuracy of the comparison methods in Fig. \ref{fig.MeTSRP} in the ascending order. From Fig. \ref{fig.MeTSRP}, we observe an interesting phenomenon that, although DICD considers more information than JDA and BDA, such as conditional distribution alignment and discriminative learning, JDA and BDA can still yield better result than DICD by generating more accurate pseudo labels. This phenomenon indicates that the accuracy of the pseudo labels has an important impact on performance. It also indicates that the accuracy of the pseudo labels can be improved by TSRP. The result also shows that TSRP can help learn better domain invariance features.

\renewcommand\arraystretch{1}
\begin{table*}[t]
\centering
\caption{Classification accuracy (\%) on the Office+Caltech-256 (decaf6 features) tasks, where A = AMAZON, C = CALTECH, D = DSLR AND W = WEBCAM}
\label{tab:decaf}
\scalebox{0.8}{
\begin{tabular}{|c||c|c|c|c|c|c|c|c|}
\hline
\diagbox{Tasks}{Methods} & NN    & NN+TSRP & JDA    & JDA+TSRP & BDA    & BDA+TSRP & DICD   & DICD+TSRP \\ \hline
C $\rightarrow$ A ($\mathrm{DeCAF_6}$)     & 85.70 & $\mathbf{88.31}$   & 89.77  & $\mathbf{92.38}$    & 90.61  & $\mathbf{92.38}$    & 91.02  & $\mathbf{92.38}$     \\ \hline
C $\rightarrow$ W ($\mathrm{DeCAF_6}$)     & 66.10 & $\mathbf{80.68}$   & 83.73  & $\mathbf{88.14}$    & 84.07  & $\mathbf{89.83}$    & 92.20  & $\mathbf{94.24}$     \\ \hline
C $\rightarrow$ D ($\mathrm{DeCAF_6}$)     & 74.52 & $\mathbf{83.44}$   & 86.62  & $\mathbf{90.45}$    & 87.90  & $\mathbf{90.45}$    & 93.63  & $\mathbf{94.90}$     \\ \hline
A $\rightarrow$ C ($\mathrm{DeCAF_6}$)     & 70.35 & $\mathbf{73.11}$   & 82.28  & $\mathbf{84.24 }$   & 83.17  & $\mathbf{84.51}$    & 86.02  & $\mathbf{87.89}$     \\ \hline
A $\rightarrow$ W ($\mathrm{DeCAF_6}$)     & 57.29 & $\mathbf{62.37}$   & 78.64  & $\mathbf{87.80}$    & 78.64  & $\mathbf{88.47}$    & 81.36  & $\mathbf{89.49}$     \\ \hline
A $\rightarrow$ D ($\mathrm{DeCAF_6}$)     & 64.97 & $\mathbf{70.06}$   & 80.25  & $\mathbf{85.35}$    & 84.71  & $\mathbf{90.45}$    & 83.44  & $\mathbf{92.36}$     \\ \hline
W $\rightarrow$ C ($\mathrm{DeCAF_6}$)     & 60.37 & $\mathbf{60.37}$   & 83.53  & 82.90               & 83.53  & $\mathbf{83.53}$    & 83.97  & $\mathbf{87.09}$     \\ \hline
W $\rightarrow$ A ($\mathrm{DeCAF_6}$)     & 62.53 & $\mathbf{66.91}$   & 90.19  & $\mathbf{91.23}$    & 90.50  & $\mathbf{91.44}$    & 89.67  & $\mathbf{90.40}$     \\ \hline
W $\rightarrow$ D ($\mathrm{DeCAF_6}$)     & 98.73 & $\mathbf{98.73}$   & 100.00 & $\mathbf{100.00}$   & 100.00 & $\mathbf{100.00}$   & 100.00 & $\mathbf{100.00}$    \\ \hline
D $\rightarrow$ C ($\mathrm{DeCAF_6}$)     & 52.09 & 49.24              & 85.13  & $\mathbf{86.82}$    & 85.22  & $\mathbf{86.82}$    & 86.11  & $\mathbf{88.33}$    \\ \hline
D $\rightarrow$ A ($\mathrm{DeCAF_6}$)     & 62.73 & $\mathbf{64.61}$   & 91.44  & $\mathbf{92.59}$    & 91.54  & $\mathbf{92.69}$    & 92.17  & $\mathbf{93.63}$     \\ \hline
D $\rightarrow$ W ($\mathrm{DeCAF_6}$)     & 89.15 & $\mathbf{96.95}$   & 98.98  & $\mathbf{98.98}$    & 98.98  & $\mathbf{99.32}$    & 98.98  & $\mathbf{98.98}$     \\ \hline
Average accuracy                          & 70.38 & $\mathbf{74.57}$   & 87.55  & $\mathbf{90.07}$    & 88.24  & $\mathbf{90.82}$    & 89.88  & $\mathbf{92.47}$     \\ \hline
Average   improvement                   &       & $\mathbf{4.19 }$   &        & $\mathbf{2.52}$     &        & $\mathbf{2.58}$     &        & $\mathbf{2.59}$      \\ \hline
Relative   improvement                  &       & $\mathbf{14.14}$   &        & $\mathbf{20.24}$    &        & $\mathbf{21.98}$    &        & $\mathbf{25.59}$    \\ \hline
\end{tabular}
}
\end{table*}

\subsection{Analysis}

The results in Tables \ref{tab:PIE} to \ref{tab:decaf} illustrate the effectiveness and generalization of TSRP, which in turn demonstrates that exploring the intra-class similarity of the target domain can remedy pseudo labels well. In this subsection, we conducted several analytical experiments to further verify the effectiveness of TSRP.

\subsubsection{Effect of UTSP on performance}
\begin{figure*}[t]
\centering
\subfigure[Validation with JDA]{
\begin{minipage}[t]{0.3\linewidth}
\centering
\includegraphics[width=1\textwidth]{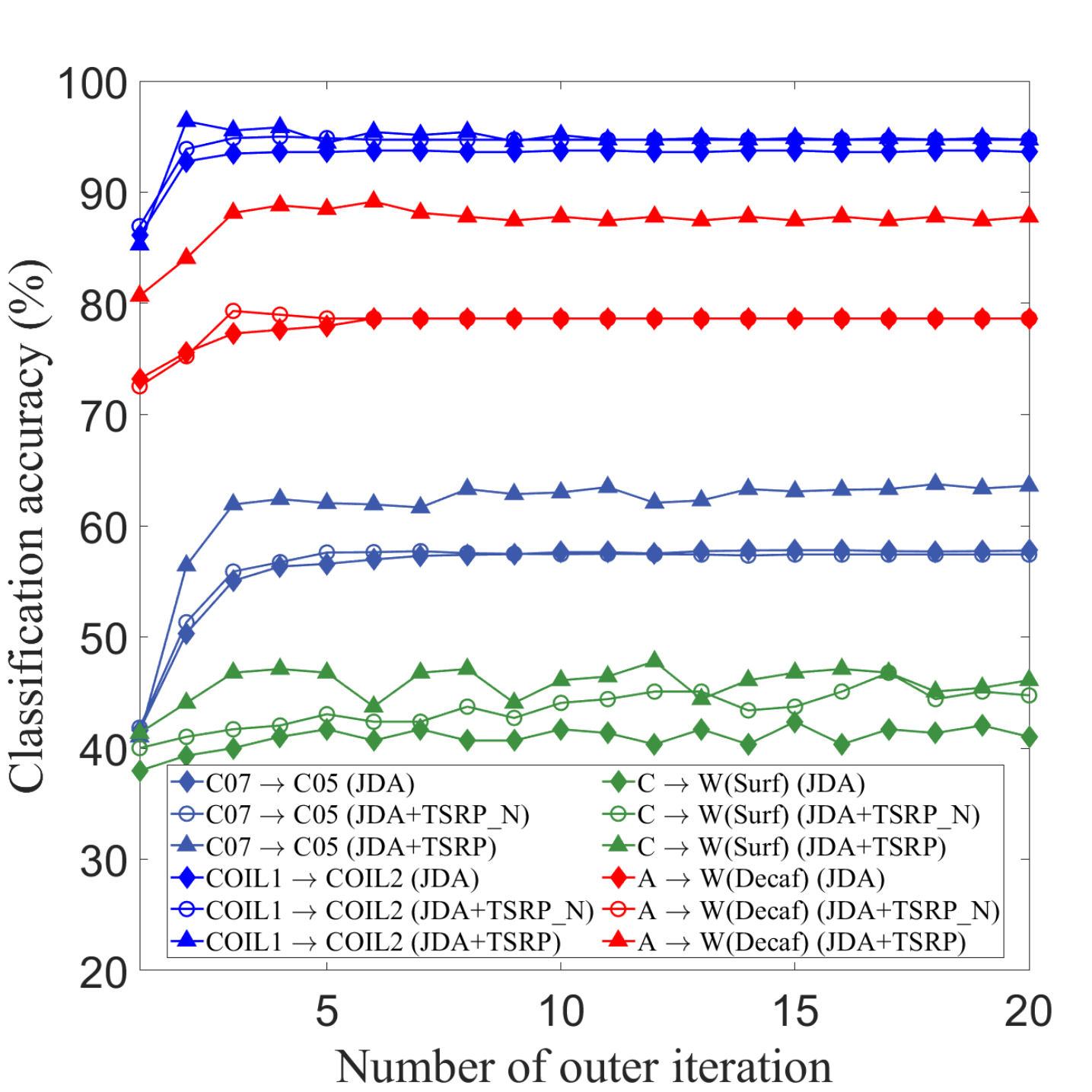}
\end{minipage}%
}%
\subfigure[Validation with BDA]{
\begin{minipage}[t]{0.3\linewidth}
\centering
\includegraphics[width=1\textwidth]{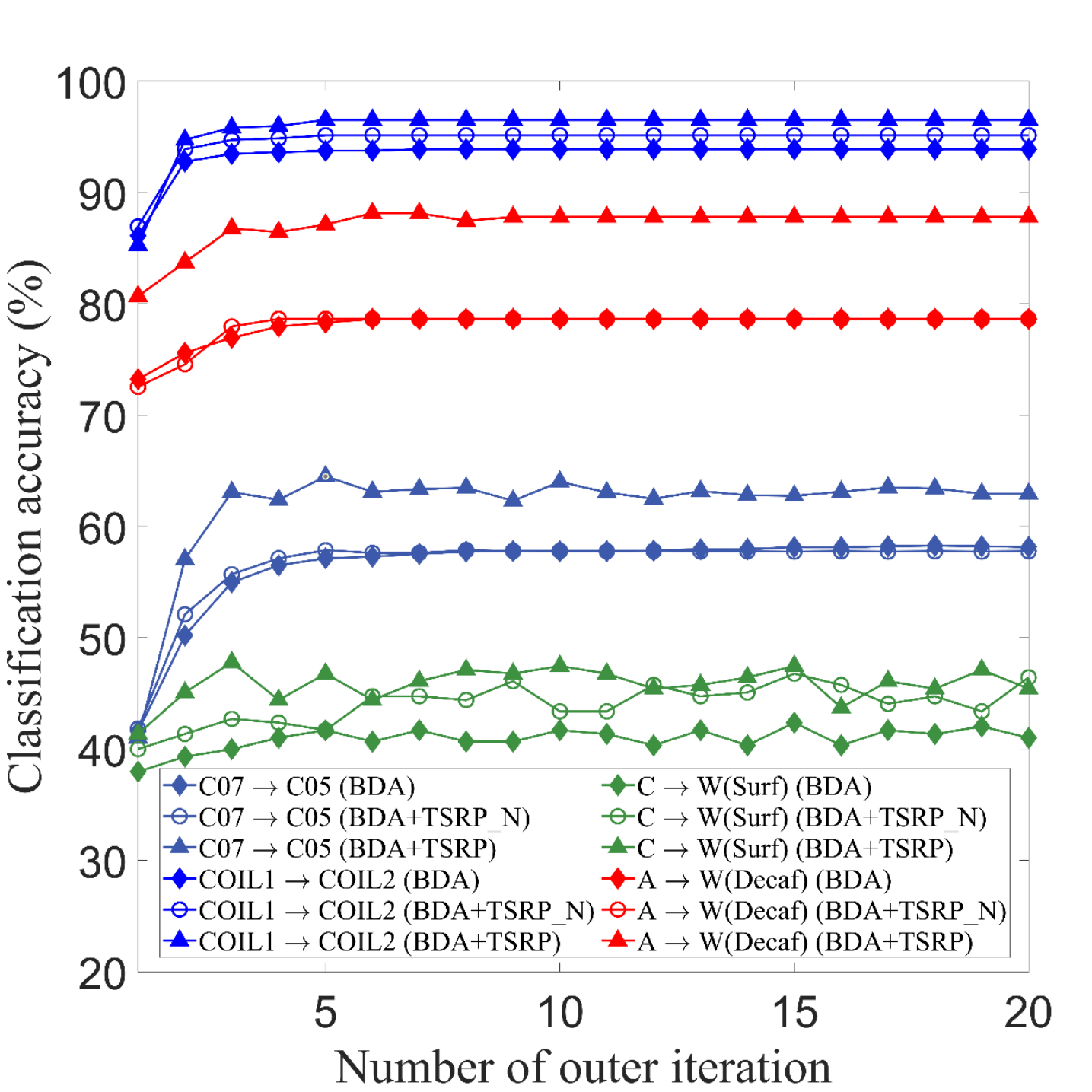}
\end{minipage}%
}
\subfigure[Validation with DICD]{
\begin{minipage}[t]{0.3\linewidth}
\centering
\includegraphics[width=1\textwidth]{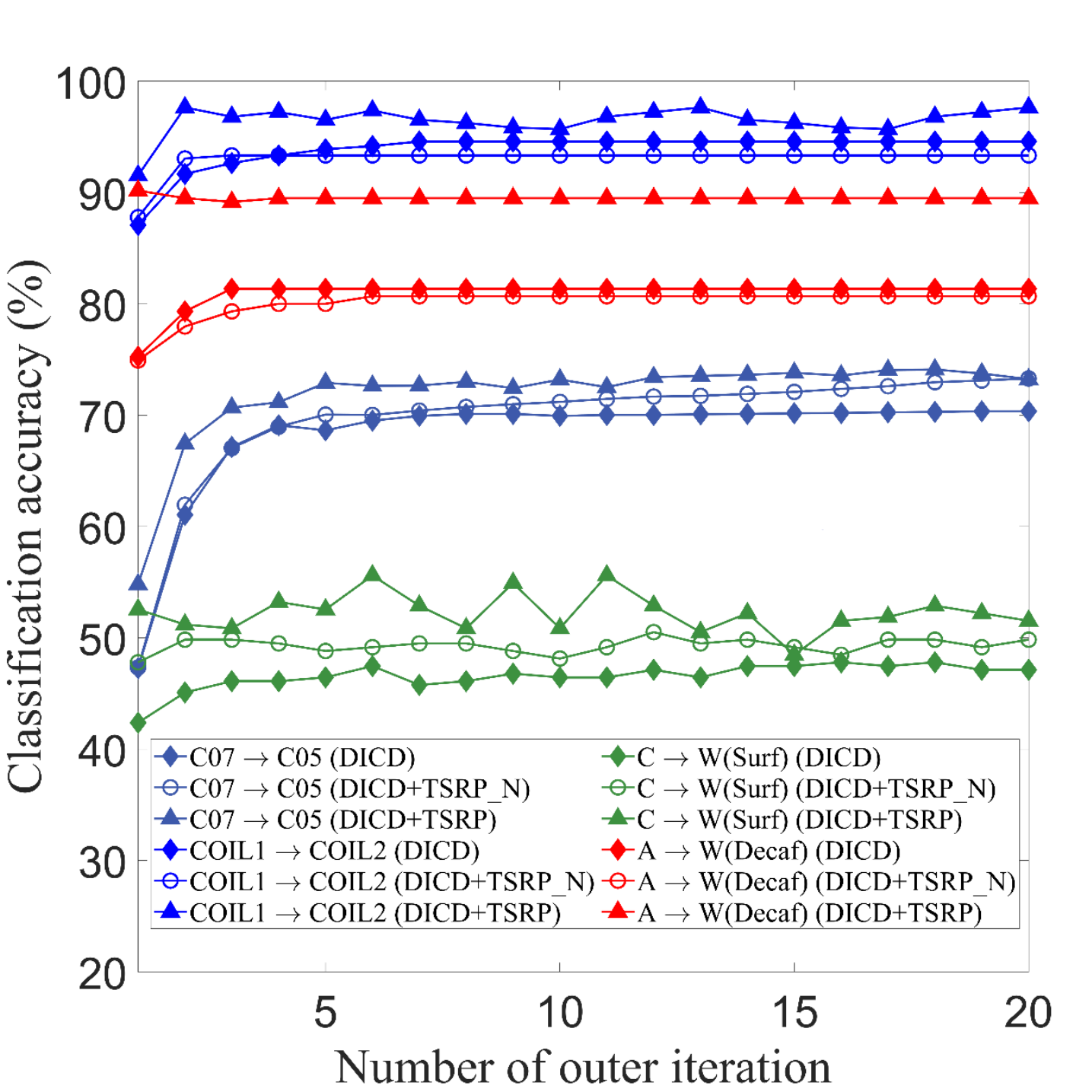}
\end{minipage}%
}%
\centering
\caption{Effect of UTSP on the performance of the Office+Caltech-256 datasets with respect to the optimization iterations. Different colors represent different domain adaptation tasks. The term ``TSRP\_N'' denotes that UTSP does not contain the \textit{spanning tree} step. }
\label{fig.UTSP}
\end{figure*}

To demonstrate the effectiveness of UTSP in picking the pseudo labels with high confidence, we compared the performance of TSRP with the proposed UTSP and a variant of UTSP that does not use the spanning tree, denoted as UTSP\_N. The result is shown in Fig. \ref{fig.UTSP}. From the figure, we can see that the TSRP without the spanning tree can still lead to a small improvement, while adding the spanning tree into TSRP can produce significantly better results. Because the main job of UTSP is to select highly confident pseudo labels, the improved performance not only supports the effectiveness of UTSP, but also reflects the effectiveness of the selected highly confident pseudo labels in promoting the alignment of the domains.

\subsubsection{Effect of the highly confident pseudo labels on the classifier}

\begin{figure*}[!ht]
\centering
\subfigure[Validation with JDA]{
\begin{minipage}[t]{0.3\linewidth}
\centering
\includegraphics[width=1\textwidth]{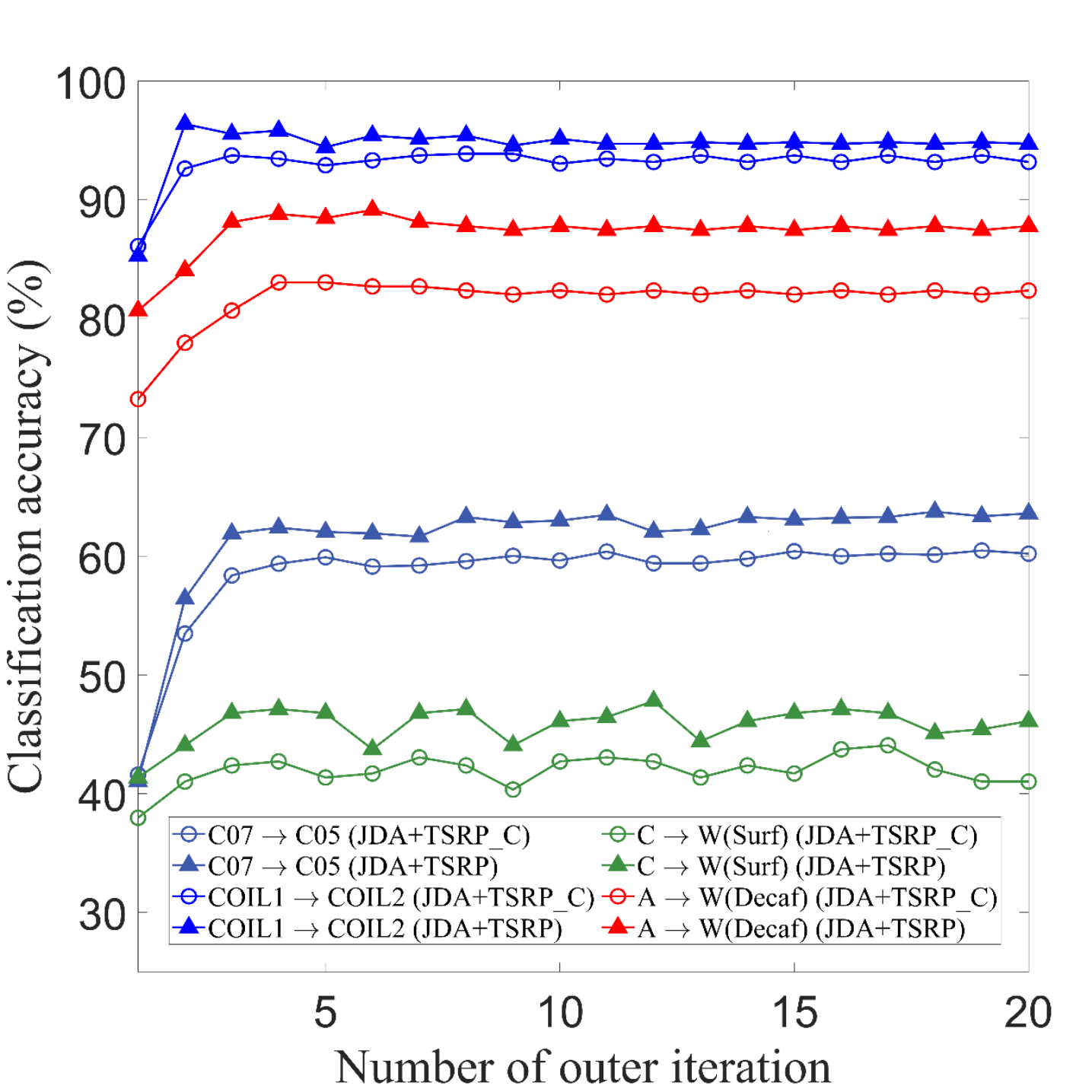}
\end{minipage}%
}%
\subfigure[Validation with BDA]{
\begin{minipage}[t]{0.3\linewidth}
\centering
\includegraphics[width=1\textwidth]{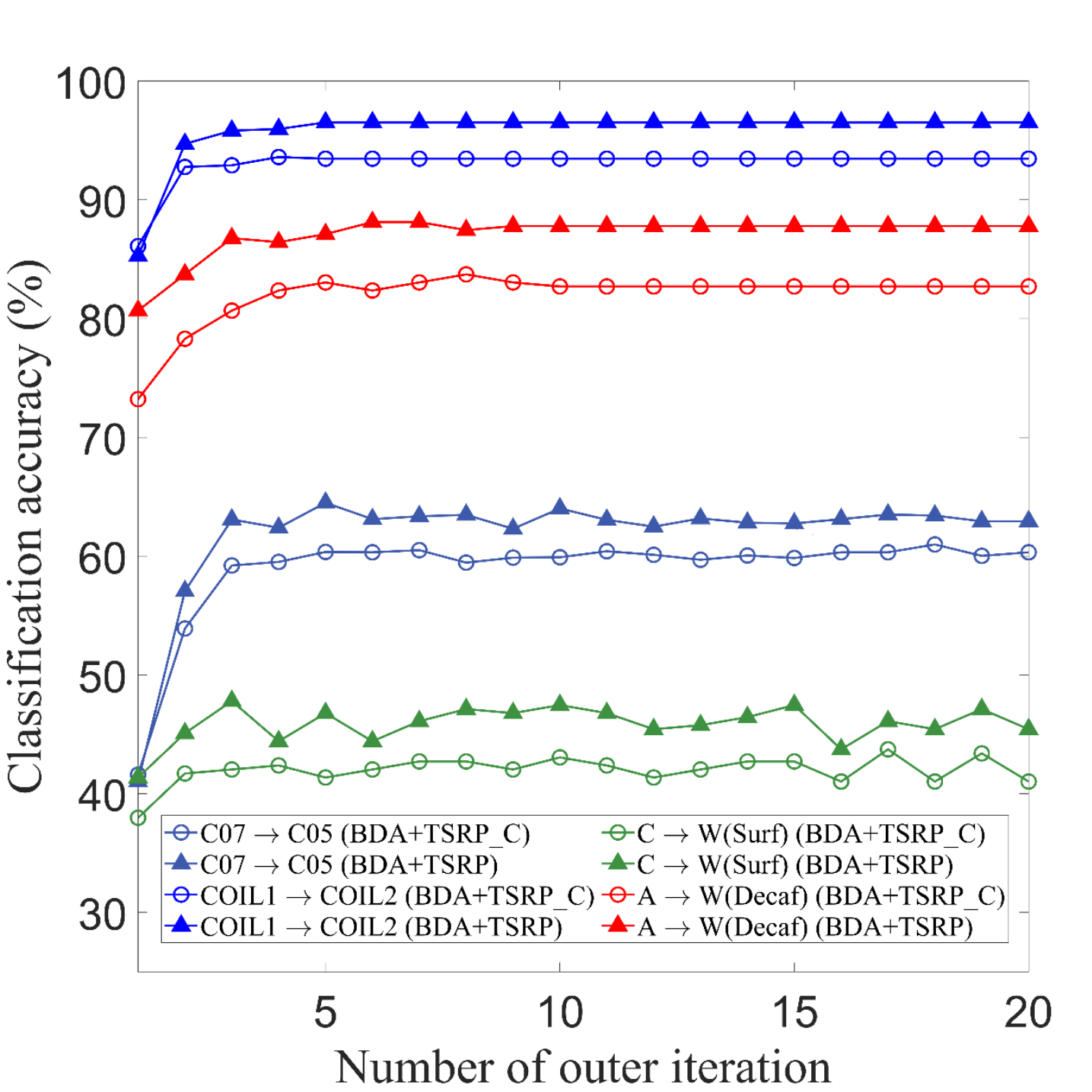}
\end{minipage}%
}
\subfigure[Validation with DICD]{
\begin{minipage}[t]{0.3\linewidth}
\centering
\includegraphics[width=1\textwidth]{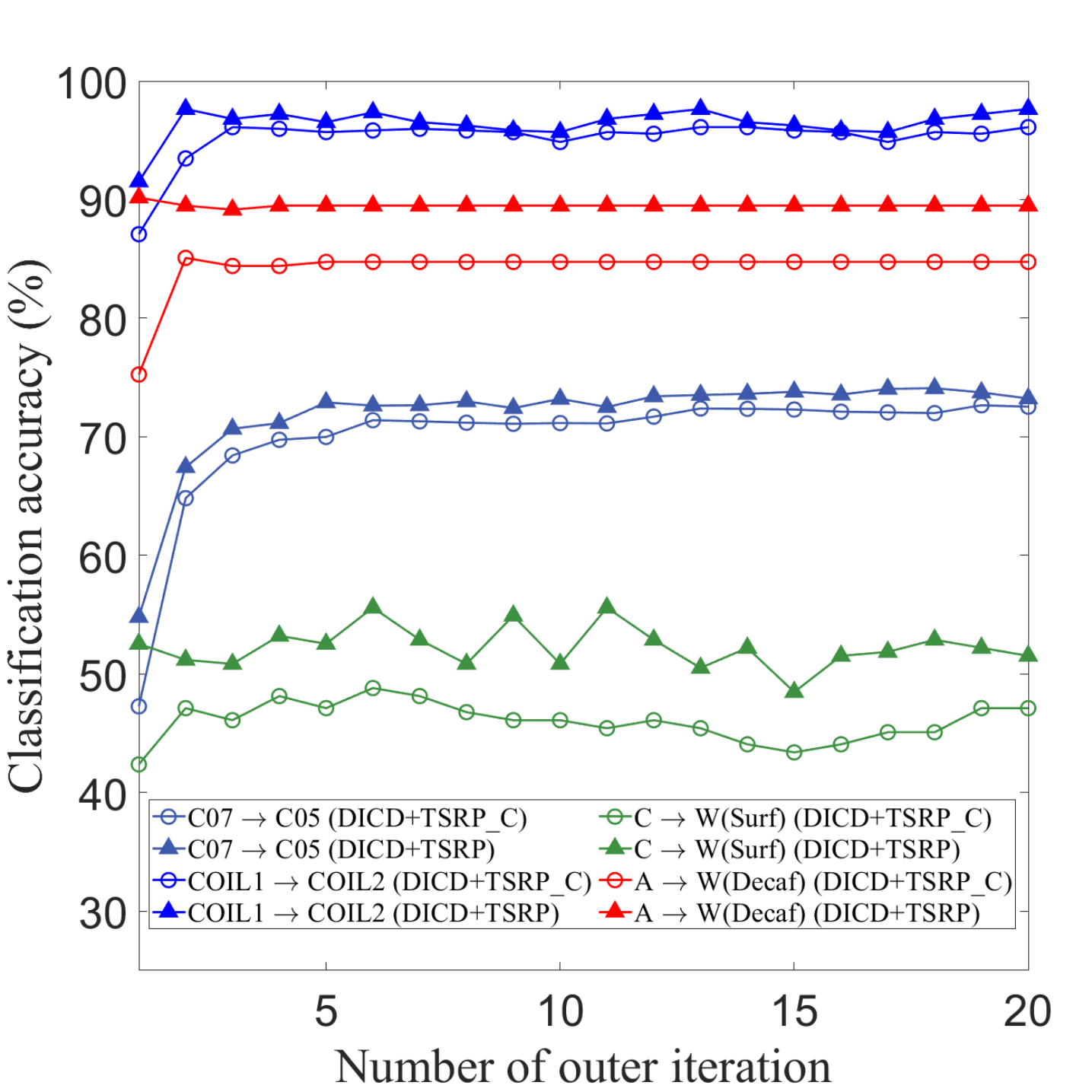}
\end{minipage}%
}%
\centering
\caption{Effect of the strong classifier in TSRP on the performance of the Office+Caltech-256 datasets with respect to the optimization iterations. The term ``TSRP\_C'' means that TSRP adopts the original classifier trained with the source data only, instead of the strong classifier.}
\label{fig.weak}
\end{figure*}

\renewcommand\arraystretch{1}
\begin{table*}[t]
\centering
\caption{Accuracy comparison of the proposed DICD+TSRP with representative UDA methods on the Office+Caltech-256 (Decaf6 features) dataset, where A = Amazon, C = Caltech, D = DSLR, and W = WEBCAM. The best result is marked in bold. The runner-up result is underlined. }\label{tab:DECAF2}
\scalebox{0.8}{
\begin{tabular}{|c||c|c|c|c|c|c|c|c|c|}
\hline \diagbox{Tasks}{Methods} & PCA & GFK & TCA & TJM & DTSL & CDML & RTML & DICD & DICD+TSRP \\
\hline $\mathrm{C} \rightarrow \mathrm{A}\left(\mathrm{DeCAF}_{6}\right)$ & $88.10$ & $87.79$ & $89.46$ & $89.46$ & $\underline{91.54}$ & $86.24$ & $90.62$ & $91.02$ & $\mathbf{92.38}$ \\
\hline $\mathrm{C} \rightarrow \mathrm{W}\left(\mathrm{DeCAF}_{6}\right)$ & $74.24$ & $70.17$ & $78.64$ & $78.98$ & $76.61$ & $77.82$ & $85.38$ & $\underline{92.20}$ & $\mathbf{94.24}$ \\
\hline $\mathrm{C} \rightarrow \mathrm{D}\left(\mathrm{DeCAF}_{6}\right)$ & $83.44$ & $88.54$ & $81.53$ & $85.35$ & $87.90$ & $83.74$ & $89.32$ & $\underline{93.6 3}$ & $\mathbf{94.90}$ \\
\hline $\mathrm{A} \rightarrow \mathrm{C}\left(\mathrm{DeCAF}_{6}\right)$ & $73.02$ & $75.78$ & $79.61$ & $79.07$ & $85.75$ & $79.54$ & $\underline{86.43}$ & $86.02$ & $\mathbf{87.89}$ \\
\hline $\mathrm{A} \rightarrow \mathrm{W}\left(\mathrm{DeCAF}_{6}\right)$ & $57.63$ & $76.95$ & $73.22$ & $76.95$ & $73.56$ & $76.27$ & $80.26$ & $\underline{81.36}$ & $\mathbf{89.49}$ \\
\hline $\mathrm{A} \rightarrow \mathrm{D}\left(\mathrm{DeCAF}_{6}\right)$ & $70.06$ & $84.08$ & $84.71$ & $\mathbf{8 5 . 3 5}$ & $82.17$ & $81.35$ & $\underline{84.36}$ & $83.44$ & $\mathbf{92.36}$ \\
\hline $\mathrm{W} \rightarrow \mathrm{C}\left(\mathrm{DeCAF}_{6}\right)$ & $63.58$ & $75.07$ & $78.09$ & $76.49$ & $72.75$ & $77.64$ & $83.13$ & $\underline{83.97}$ & $\mathbf{87.09}$ \\
\hline $\mathrm{W} \rightarrow \mathrm{A}\left(\mathrm{DeCAF}_{6}\right)$ & $70.15$ & $82.88$ & $83.30$ & $86.74$ & $75.47$ & $86.29$ & $\mathbf{91.37}$ & $89.67$ & $\underline{90.40}$ \\
\hline $\mathrm{W} \rightarrow \mathrm{D}\left(\mathrm{DeCAF}_{6}\right)$ & $\mathbf{1 0 0 . 0 0}$ & $\mathbf{1 0 0 . 0 0}$ & $\mathbf{1 0 0 . 0 0}$ & $\mathbf{1 0 0 . 0 0}$ & $\mathbf{1 0 0 . 0 0}$ & $98.42$ & $\mathbf{1 0 0 . 0 0}$ & $\mathbf{1 0 0 . 0 0}$ & $\mathbf{1 0 0 . 0 0}$ \\
\hline $\mathrm{D} \rightarrow \mathrm{C}\left(\mathrm{DeCAF}_{6}\right)$ & $57.88$ & $73.11$ & $79.70$ & $78.63$ & $75.24$ & $78.56$ & $85.72$ & $\underline{86.11}$ & $\mathbf{88.33}$ \\
\hline $\mathrm{D} \rightarrow \mathrm{A}\left(\mathrm{DeCAF}_{6}\right)$ & $68.16$ & $85.18$ & $88.52$ & $89.77$ & $84.97$ & $89.47$ & $91.86$ & $\underline{92.17}$ & $\mathbf{93.63}$ \\
\hline $\mathrm{D} \rightarrow \mathrm{W}\left(\mathrm{DeCAF}_{6}\right)$ & $88.14$ & $90.85$ & $\underline{98.98}$ & $97.97$ & $\mathbf{99.32}$ & $96.38$ & $\underline{98.98}$ & $\underline{98.98}$ & $\underline{98.98}$ \\
\hline Average accuracy & $74.53$ & $82.53$ & $84.65$ & $85.40$ & $83.77$ & $84.31$ & $88.95$ & $\underline{89.88}$ & $\mathbf{92.47}$ \\
\hline
\end{tabular}
}
\end{table*}
\begin{figure*}[!ht]
\centering
\subfigure[JDA]{
\begin{minipage}[t]{0.20\linewidth}
\centering
\includegraphics[width=1\textwidth]{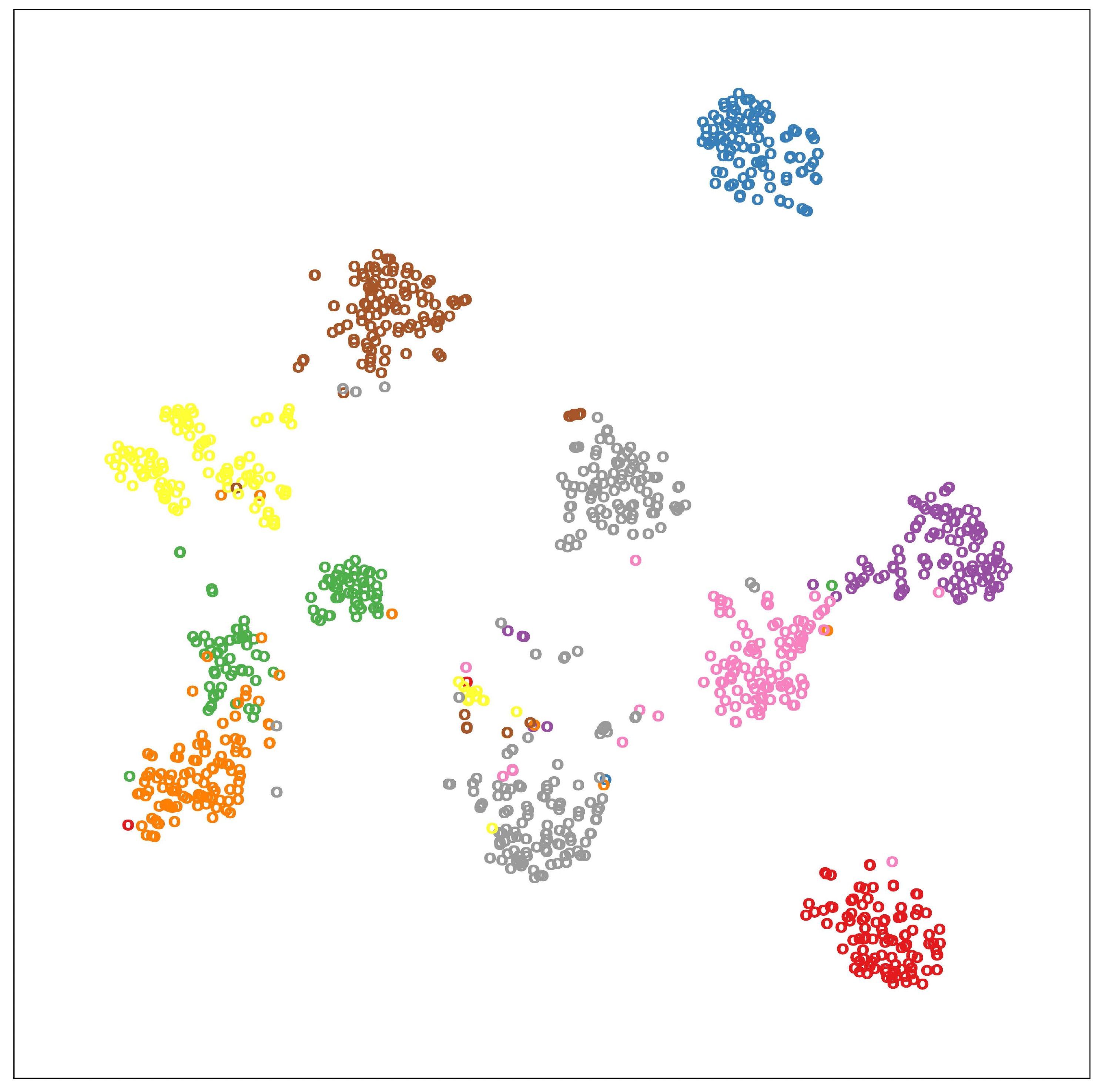}
\end{minipage}%
}%
\subfigure[JDA+TSRP]{
\begin{minipage}[t]{0.20\linewidth}
\centering
\includegraphics[width=1\textwidth]{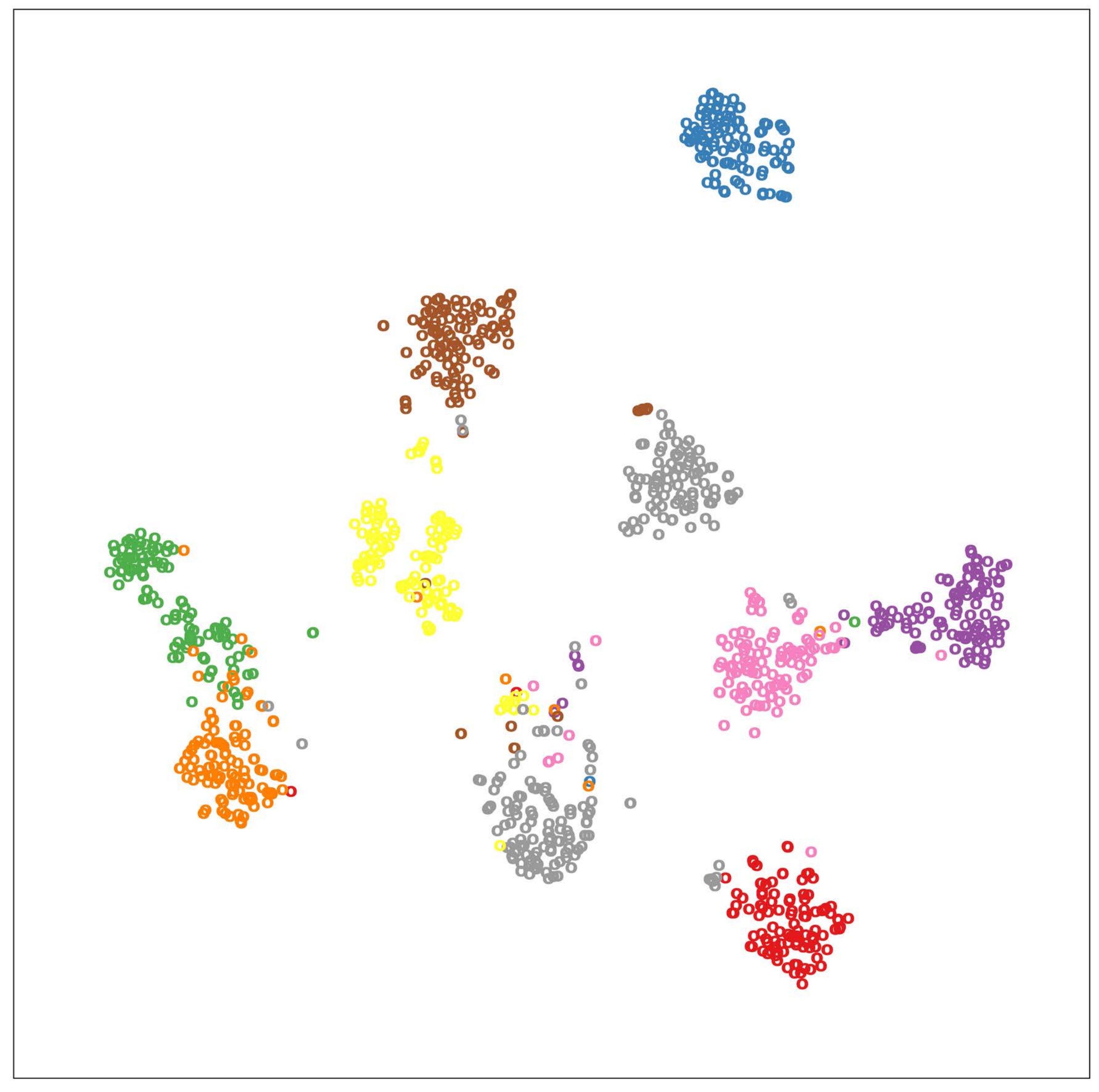}
\end{minipage}%
}\hspace{-1mm}
\subfigure[DICD]{
\begin{minipage}[t]{0.20\linewidth}
\centering
\includegraphics[width=1\textwidth]{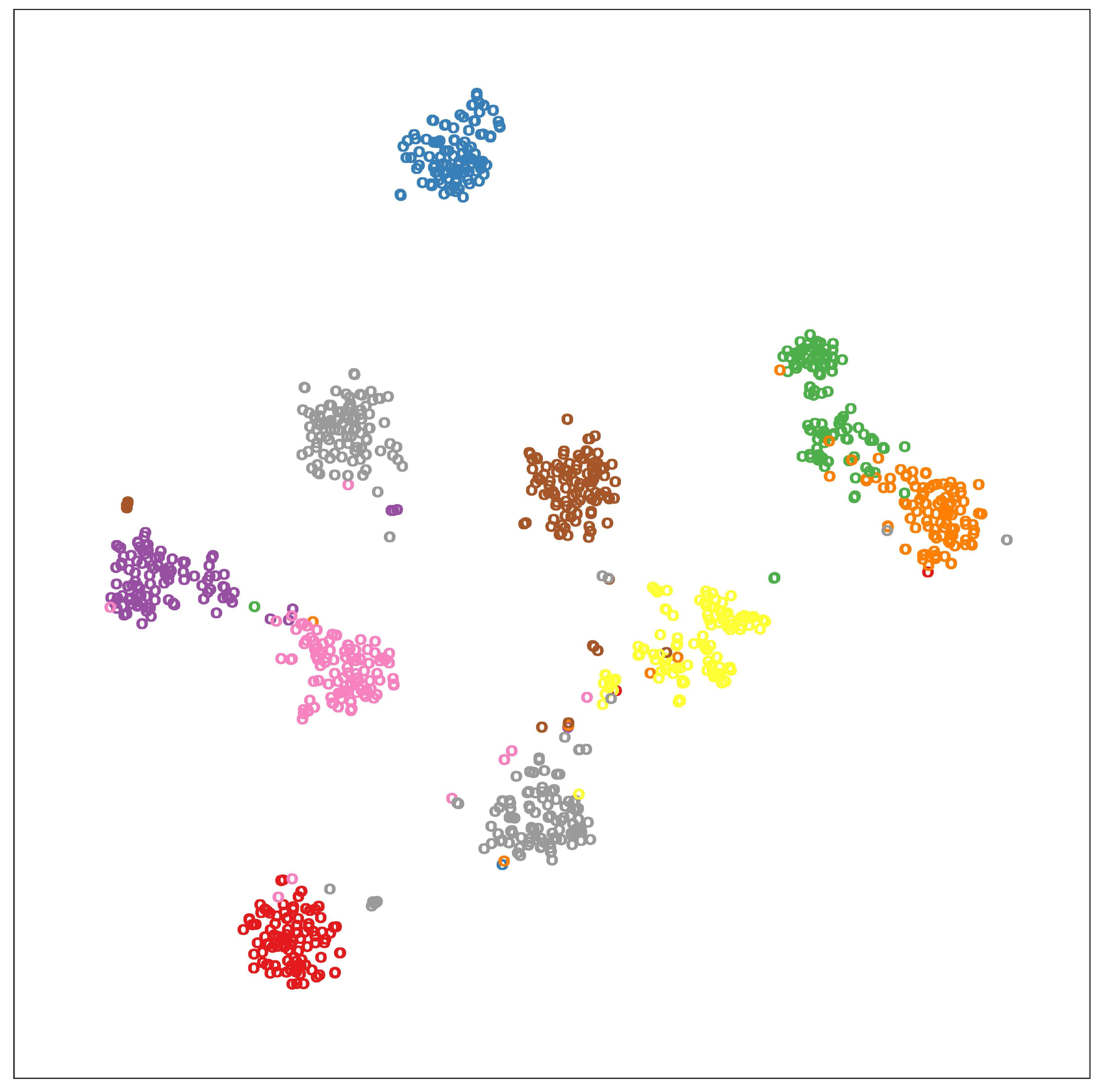}
\end{minipage}%
}%
\subfigure[DICD+TSRP]{
\begin{minipage}[t]{0.20\linewidth}
\centering
\includegraphics[width=1\textwidth]{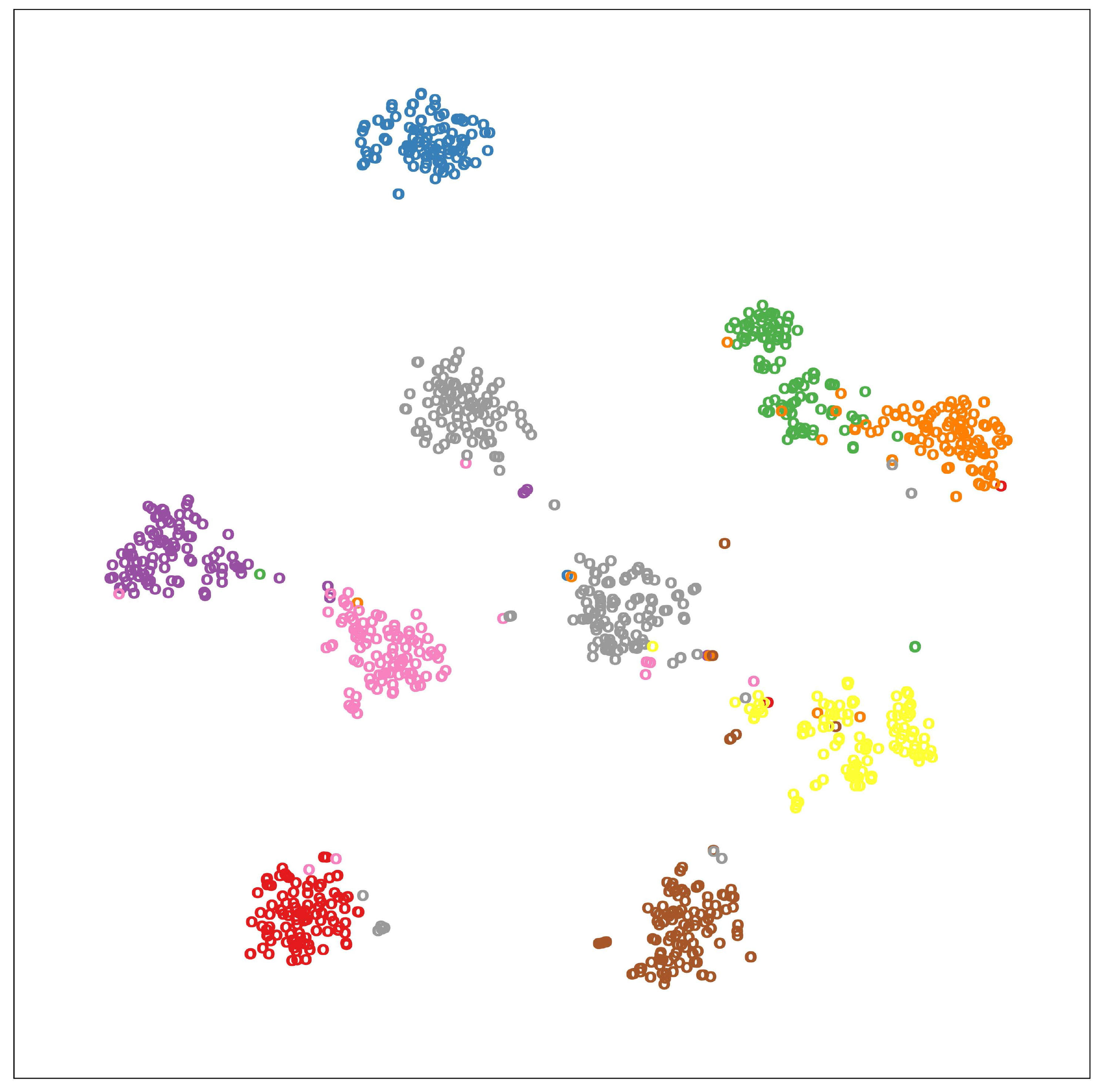}
\end{minipage}%
}
\caption{Visualization of the features produced by the comparison methods. Different colors represent different categories.}
\label{fig.tsne}
\end{figure*}
\begin{figure}[t]
    \centering
    \includegraphics[width=0.65\textwidth]{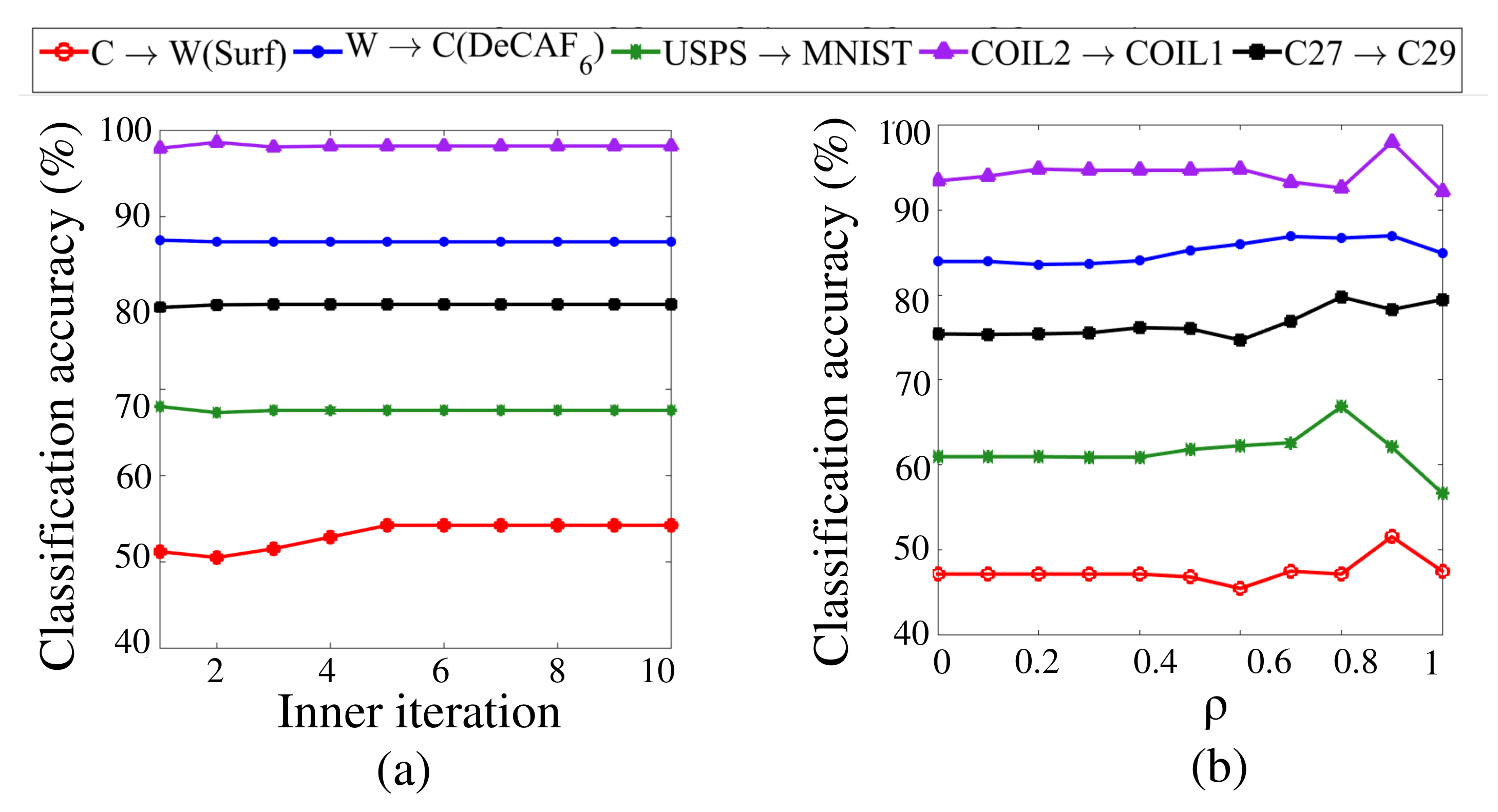}\\
    \caption{Effect of the hyperparameters on five domain adaptation tasks. Different colors represent different domain adaptation tasks.}\label{fig.can}
\end{figure}

To demonstrate the effectiveness of the selected highly confident pseudo labels in improving the performance of the classifier, we compared the classifier (denoted as ``strong classifier'') with the one that is trained with the source samples (denoted as ``original classifier'') only. The result is shown in Fig. \ref{fig.weak}. From the figure, we see that the performance of the
strong classifier is much better than the original classifier. It indicates that TSRP can help learn good domain-invariant and discriminative features, and in turn, the features can help TSRP refine the pseudo labels, which is a co-promotion process.

From Fig. \ref{fig.weak}, we can also find that, no matter how many iterations are conducted, the baseline methods without TSRP are upper-bounded due to the low accuracy of the pseudo labels. On the other side, the proposed methods with TSRP can break through such limit.
\subsubsection{Effects of hyper-parameters on performance}
Our approach consists of two hyper-parameters: The trust parameter $\rho$ and the number of inner iterations $IT$. Here we take DICD+TSRP as an example to study how the two hyperparameters affect the performance. The experiment was conducted on the tasks $\mathrm{C27}\rightarrow\mathrm{C29}$, $\mathrm{C}\rightarrow\mathrm{W}$ (SURF), $\mathrm{W}\rightarrow\mathrm{C}$ (DeCAF6), $\mathrm{COIL2}\rightarrow\mathrm{COIL1}$ and $\mathrm{USPS}\rightarrow\mathrm{MNIST}$. The results are reported in Fig. \ref{fig.can}. Specifically, Fig. \ref{fig.can}a shows the effect of $IT$
when $\rho$ was set to $0.9$ and $IT$ was selected from $\{1,2, \ldots, 10\}$. From the figure, we see obviously that our approach is not sensitive to $IT$. Figure \ref{fig.can}b shows the effect of $\rho$ when $IT = 4$ and $\rho$ was selected from $\{0, 0.1, 0.2, \ldots, 1\}$. The result indicates that, although the the proposed method is relatively sensitive to the trust parameter $\rho$ for each single task, it reaches the best performance on all tasks when $\rho\approx 0.8$.

\subsubsection{Visualization}
To demonstrate how the proposed method improves the alignment of the source and target domains, we visualize the projected representations of JDA, JDA+TSRP, DICD and DICD+TSRP on the task $A \rightarrow D$ (DeCAF6 features) in Fig. \ref{fig.tsne}. Comparing Figs. \ref{fig.tsne}a and \ref{fig.tsne}b, we can see that TSRP+JDA has a better aligned conditional distribution than JDA. Comparing Figs. \ref{fig.tsne}c and \ref{fig.tsne}d, we observe similar phenomenon, which supports our claim on the advantage of TSRP.

\subsection{Comparison with other domain adaptation methods}
To further illustrate the effectiveness of TSRP, we compared DICD+TSRP with several other UDA methods, which are GFK \cite{geodestic}, TCA \cite{TCA}, TJM \cite{TJM}, TSL \cite{TSL}, DTSL \cite{DTSL}, CDML \cite{CDML}, RTML \cite{RTML}, and DICD \cite{DICD}. We also compared with a standard machine learning methods, i.e., PCA \cite{PCA}. In order to make a fair comparison, the results of the comparison methods are from their public codes or the original papers.

Table \ref{tab:DECAF2} shows the classification accuracy of the comparison methods on the Office+Caltech-256 (DECAF6 features) dataset. From the table, we can see that DICD+TSRP achieves the best performance in 10 out of 12 tasks, and ranks the second in the other 2 tasks. The results on the other datasets are listed in the supplementary material. The experimental conclusion is similar to that on the Office+Caltech-256 (DECAF6 features) dataset.

\section{Conclusion}\label{sec:conclusion}
In this paper, we have proposed to {use target domain intra-class similarity to remedy pseudo labels} (TSRP) for improving the accuracy of the coarse pseudo labels that are generated from a conventional UDA method, which in turn improves the discriminant ability of the learned representation of the UDA. Specifically, TSRP first exploits the intra-class similarity and spanning trees to pick samples with high confident pseudo labels. Then, it trains a strong classifier with both the source samples and the target samples whose pseudo labels are highly-confident. Finally, it uses the strong classifier to remedy the pseudo labels of the target samples with low-confident pseudo labels. Experimental results on extensive visual cross-domain tasks have shown that applying TSRP to conventional UDA methods can improve the accuracy of the pseudo labels and further lead to more discriminative and domain invariant features than the conventional UDA baselines.
\section*{References}
\normalem
\bibliographystyle{elsarticle-num}
\bibliography{reference}


\pagebreak

\begin{table*}[t]
\centering
\caption{[supplementary material] Accuracy comparison (\%) of the proposed DICD+TSRP with representative UDA methods on the CMU-PIE dataset.}
\label{tab:PIE2}
\scalebox{0.8}{
\begin{tabular}{|c||c|c|c|c|c|c|c|c||c|}
\hline \diagbox{Tasks}{Methods} & PCA & GFK & TCA & TSL & DTSL & CDML & RTML & DICD & DICD+TSRP \\
\hline $\mathrm{C} 05 \rightarrow$ C07 & $24.80$ & $26.15$ & $40.76$ & $44.08$ & $65.87$ & $53.22$ & $60.12$ & $\underline{72.99}$ & $\mathbf{74.52}$ \\
\hline $\mathrm{C} 05 \rightarrow \mathrm{C} 09$ & $25.18$ & $27.27$ & $41.79$ & $47.49$ & $64.09$ & $53.12$ & $55.21$ & $\underline{72.00}$ & $\mathbf{76.16}$ \\
\hline $\mathrm{C} 05 \rightarrow \mathrm{C} 27$ & $29.26$ & $31.15$ & $59.63$ & $62.78$ & $82.03$ & $80.12$ & $85.19$ & $\underline{92.22}$ & $\mathbf{96.52}$ \\
\hline $\mathrm{C} 05 \rightarrow \mathrm{C} 29$ & $16.30$ & $17.59$ & $29.35$ & $36.15$ & $54.90$ & $48.23$ & $52.98$ & $\underline{66.85}$ & $\mathbf{71.32}$ \\
\hline $\mathrm{C} 07 \rightarrow \mathrm{C} 05$ & $24.22$ & $25.24$ & $41.81$ & $46.28$ & $45.04$ & $52.39$ & $58.13$ & $\underline{69.93}$ & $\mathbf{73.20}$ \\
\hline $\mathrm{C} 07 \rightarrow \mathrm{C} 09$ & $45.53$ & $47.37$ & $51.47$ & 57.60 & 53.49 & 54.23 & 63.92 & $\underline{65.87}$ & $\mathbf{67.83}$ \\
\hline $\mathrm{C} 07 \rightarrow \mathrm{C} 27$ & 53.35 & 54.25 & 64.73 & 71.43 & 71.43 & 68.36 & 76.16 & $\underline{85.25}$ & $\mathbf{86.42}$ \\
\hline $\mathrm{C} 07 \rightarrow \mathrm{C} 29$ & 25.43 & 27.08 & 33.70 & 35.66 & 47.97 & 37.34 & 40.38 & $\underline{48.31}$ & $\mathbf{56.43}$ \\
\hline $\mathrm{C} 09 \rightarrow \mathrm{C} 05$ & 20.95 & 21.82 & 34.69 & 36.94 & 52.49 & 43.54 & 53.12 & $\underline{69.36}$ & $\mathbf{70.14}$ \\
\hline $\mathrm{C} 09 \rightarrow \mathrm{C} 07$ & 40.45 & 43.16 & 47.70 & 47.02 & 55.56 & 54.87 & 58.67 & $\underline{65.44}$ & $\mathbf{73.05}$ \\
\hline $\mathrm{C} 09 \rightarrow \mathrm{C} 27$ & 46.14 & 46.41 & 56.23 & 59.45 & 77.50 & 62.76 & 69.81 & $\underline{83.39}$ & $\mathbf{94.26}$ \\
\hline $\mathrm{C} 09 \rightarrow \mathrm{C} 29$ & 25.31 & 26.78 & 33.15 & 36.34 & 54.11 & 38.21 & 42.13 & $\underline{61.40}$ & $\mathbf{66.48}$ \\
\hline $\mathrm{C} 27 \rightarrow \mathrm{C} 05$ & 31.96 & 34.24 & 55.64 & 63.66 & 81.54 & 75.12 & 81.12 & $\underline{93.13}$ & $\mathbf{94.72}$ \\
\hline $\mathrm{C} 27 \rightarrow \mathrm{C} 07$ & 60.96 & 62.92 & 67.83 & 72.68 & 85.39 & 80.53 & 83.92 & $\underline{90.12}$ & $\mathbf{92.88}$ \\
\hline $\mathrm{C} 27 \rightarrow \mathrm{C} 09$ & 72.18 & 73.35 & 75.86 & 83.52 & 82.23 & 83.72 & $\underline{89.51}$ & 88.97 & $\mathbf{90.26}$ \\
\hline $\mathrm{C} 27 \rightarrow \mathrm{C} 29$ & 35.11 & 37.38 & 40.26 & 44.79 & 72.61 & 52.78 & 56.26 & $\underline{75.61}$ & $\mathbf{79.11}$ \\
\hline $\mathrm{C} 29 \rightarrow \mathrm{C} 05$ & 18.85 & 20.35 & 26.98 & 33.28 & $52.19$ & 27.34 & 29.11 & $\underline{62.88}$ & $\mathbf{73.68}$ \\
\hline $\mathrm{C} 29 \rightarrow \mathrm{C} 07$ & 23.39 & 24.62 & 29.90 & 34.13 & 49.41 & 30.82 & 33.28 & $\underline{57.03}$ & $\mathbf{65.81}$ \\
\hline $\mathrm{C} 29 \rightarrow \mathrm{C} 09$ & 27.21 & 28.49 & 29.90 & 36.58 & $58.45$ & 36.34 & 39.85 & $\underline{65.87}$ & $\mathbf{70.10}$ \\
\hline $\mathrm{C} 29 \rightarrow \mathrm{C} 29$ & 30.34 & 31.33 & 33.64 & 38.75 & $64.31$ & 40.61 & 47.13 & $\underline{74.77}$ & $\mathbf{77.84}$ \\
\hline Average accuracy & 33.85 & 35.35 & 44.75 & 49.43 & 63.53 & 53.68 & 58.80 & $\underline{73.09}$ & $\mathbf{77.84}$ \\
\hline
\end{tabular}
}
\end{table*}
\begin{table*}[h]
\centering
\caption{[supplementary material] Accuracy comparison (\%) of the proposed DICD+TSRP with representative UDA methods on the Office+Caltech-256 (surf features), MNIST+USPS, and COIL20 datasets.}\label{tab:SURF2}
\scalebox{0.8}{
\begin{tabular}{|c||c|c|c|c|c|c|c|c|c|}
\hline \diagbox{Tasks}{Methods} & PCA & GFK & TCA & TJM & DTSL & CDML & RTML & DICD & DICD+TSRP \\
\hline $\mathrm{C} \rightarrow \mathrm{A}$ (SURF) & $36.95$ & $41.02$ & $38.20$ & $46.76$ & $\mathbf{51.25}$ & $47.82$ & $\underline{49.26}$ & $47.29$ & $47.81$ \\
\hline $\mathrm{C} \rightarrow \mathrm{W}$ (SURF) & $32.54$ & $40.68$ & $38.64$ & $38.98$ & $38.64$ & $36.91$ & $44.72$ & $\underline{46.44}$ & $\mathbf{50.85}$ \\
\hline $\mathrm{C} \rightarrow \mathrm{D}$ (SURF) & $38.22$ & $38.85$ & $41.40$ & $44.59$ & $47.13$ & $43.93$ & $47.56$ & $\underline{49.68}$ & $\mathbf{50.96}$ \\
\hline $\mathrm{A} \rightarrow \mathrm{C}$ (SURF) & $34.73$ & $40.25$ & $37.76$ & $39.45$ & $\underline{43.37}$ & $41.72$ & $\mathbf{43.68}$ & $42.39$ & $41.76$ \\
\hline $\mathrm{A} \rightarrow \mathrm{W}$ (SURF) & $35.59$ & $38.98$ & $37.63$ & $42.03$ & $36.61$ & $38.25$ & $44.32$ & $\underline{45.08}$ & $\mathbf{49.15}$ \\
\hline $\mathrm{A} \rightarrow \mathrm{D}$ (SURF) & $27.39$ & $36.31$ & $33.12$ & $\mathbf{45.22}$ & $38.85$ & $35.92$ & $\underline{43.86}$ & $38.85$ & $42.04$ \\
\hline $\mathrm{W} \rightarrow \mathrm{C}$ (SURF) & $26.36$ & $30.72$ & $29.30$ & $30.19$ & $29.83$ & $31.14$ & $\mathbf{3 4 . 8 3}$ & $\underline{33.57}$ & $32.95$ \\
\hline $\mathrm{W} \rightarrow \mathrm{A}$ (SURF) & $31.00$ & $29.75$ & $30.06$ & $29.96$ & $\underline{34.13}$ & $32.26$ & $\mathbf{3 5 . 2 8}$ & $\underline{34.13}$ & $31.94$ \\
\hline $\mathrm{W} \rightarrow \mathrm{D}$ (SURF) & $77.07$ & $80.89$ & $87.26$ & $89.17$ & $82.80$ & $84.84$ & $\mathbf{9 1 . 0 2}$ & $\underline{89.81}$ & $\underline{89.81}$ \\
\hline $\mathrm{D} \rightarrow \mathrm{C}$ (SURF) & $29.65$ & $30.28$ & $31.70$ & $31.43$ & $30.11$ & $32.63$ & $34.58$ & $\underline{34.64}$ & $\mathbf{37.04}$ \\
\hline $\mathrm{D} \rightarrow \mathrm{A}$ (SURF) & $32.05$ & $32.05$ & $32.15$ & $32.78$ & $32.05$ & $29.87$ & $33.26$ & $\underline{34.45}$ & $\mathbf{35.28}$ \\
\hline D $\rightarrow$ W (SURF) & $75.93$ & $75.59$ & $86.10$ & $85.42$ & $72.20$ & $82.34$ & $\underline{89.68}$ & $\mathbf{91.19}$ & $\mathbf{91.19}$ \\
\hline USPS $\rightarrow$ MNIST & $44.95$ & $46.45$ & $51.05$ & $52.25$ & $55.50$ & $52.25$ & $\underline{61.82}$ & $61.50$ & $\mathbf{67.55}$ \\
\hline MNIST $\rightarrow$ USPS & $66.22$ & $67.22$ & $56.28$ & $63.28$ & $52.33$ & $63.28$ & $69.52$ & $\underline{73.28}$ & $\mathbf{73.39}$ \\
\hline COIL1 $\rightarrow$ COIL2 & $84.72$ & $72.50$ & $88.47$ & $91.53$ & $88.61$ & $88.93$ & $91.23$ & $\underline{94.58}$ & $\mathbf{95.69}$ \\
\hline COIL2 $\rightarrow$ COIL1 & $84.03$ & $74.17$ & $85.83$ & $91.81$ & $89.17$ & $87.32$ & $90.22$ & $\underline{93.47}$ & $\mathbf{98.06}$ \\
\hline Average accuracy & 47.34 & 48.48 & 50.31 & 53.43 & 51.41 & 51.84 & 56.55 & $\underline{56.90}$ & $\mathbf{58.46}$ \\
\hline
\end{tabular}
}
\end{table*}

\end{document}